\documentclass[10pt,twocolumn,letterpaper]{article}

\usepackage{iccv}
\usepackage{times}
\usepackage{epsfig}
\usepackage{graphicx}
\usepackage{amsmath}
\usepackage{amssymb}
\usepackage{bbm}
\usepackage[misc]{ifsym}
\usepackage{abstract}
\usepackage[accsupp]{axessibility}  

\usepackage[pagebackref=true,breaklinks=true,letterpaper=true,colorlinks,bookmarks=false]{hyperref}

\usepackage[capitalize]{cleveref}
\crefname{section}{Sec.}{Secs.}
\Crefname{section}{Section}{Sections}
\Crefname{table}{Table}{Tables}
\crefname{table}{Tab.}{Tabs.}

\iccvfinalcopy 

\def\iccvPaperID{5047} 

\begin{document}

\title{Two-in-One Depth: Bridging the Gap Between Monocular and Binocular Self-supervised Depth Estimation}

\author{Zhengming Zhou and Qiulei Dong \textsuperscript{\Letter}\\
State Key Laboratory of Multimodal Artificial Intelligence Systems, CASIA\\
School of Artificial Intelligence, UCAS\\
{\tt\small zhouzhengming2020@ia.ac.cn qldong@nlpr.ia.ac.cn}
}

\maketitle


\renewcommand{\thefootnote}{}
\footnotetext{\Letter \enspace corresponding author}

\begin{abstract}
  Monocular and binocular self-supervised depth estimations are two important and related tasks in computer vision, which aim to predict scene depths from single images and stereo image pairs respectively.
  In literature, the two tasks are usually tackled separately by two different kinds of models, and binocular models generally fail to predict depth from single images, while the prediction accuracy of monocular models is generally inferior to binocular models.
  In this paper, we propose a Two-in-One self-supervised depth estimation network, called TiO-Depth, which could not only compatibly handle the two tasks, but also improve the prediction accuracy.
  TiO-Depth employs a Siamese architecture and each sub-network of it could be used as a monocular depth estimation model.
  For binocular depth estimation, a Monocular Feature Matching module is proposed for incorporating the stereo knowledge between the two images, and the full TiO-Depth is used to predict depths.
  We also design a multi-stage joint-training strategy for improving the performances of TiO-Depth in both two tasks by combining the relative advantages of them.
  Experimental results on the KITTI, Cityscapes, and DDAD datasets demonstrate that TiO-Depth outperforms both the monocular and binocular state-of-the-art methods in most cases, and further verify the feasibility of a two-in-one network for monocular and binocular depth estimation.
  The code is available at https://github.com/ZM-Zhou/TiO-Depth\_pytorch.
  \end{abstract}

\section{Introduction}
\label{sec:intro}

\begin{figure}[]
  \footnotesize
  \begin{center}
  \centerline{\includegraphics[width=0.88\linewidth]{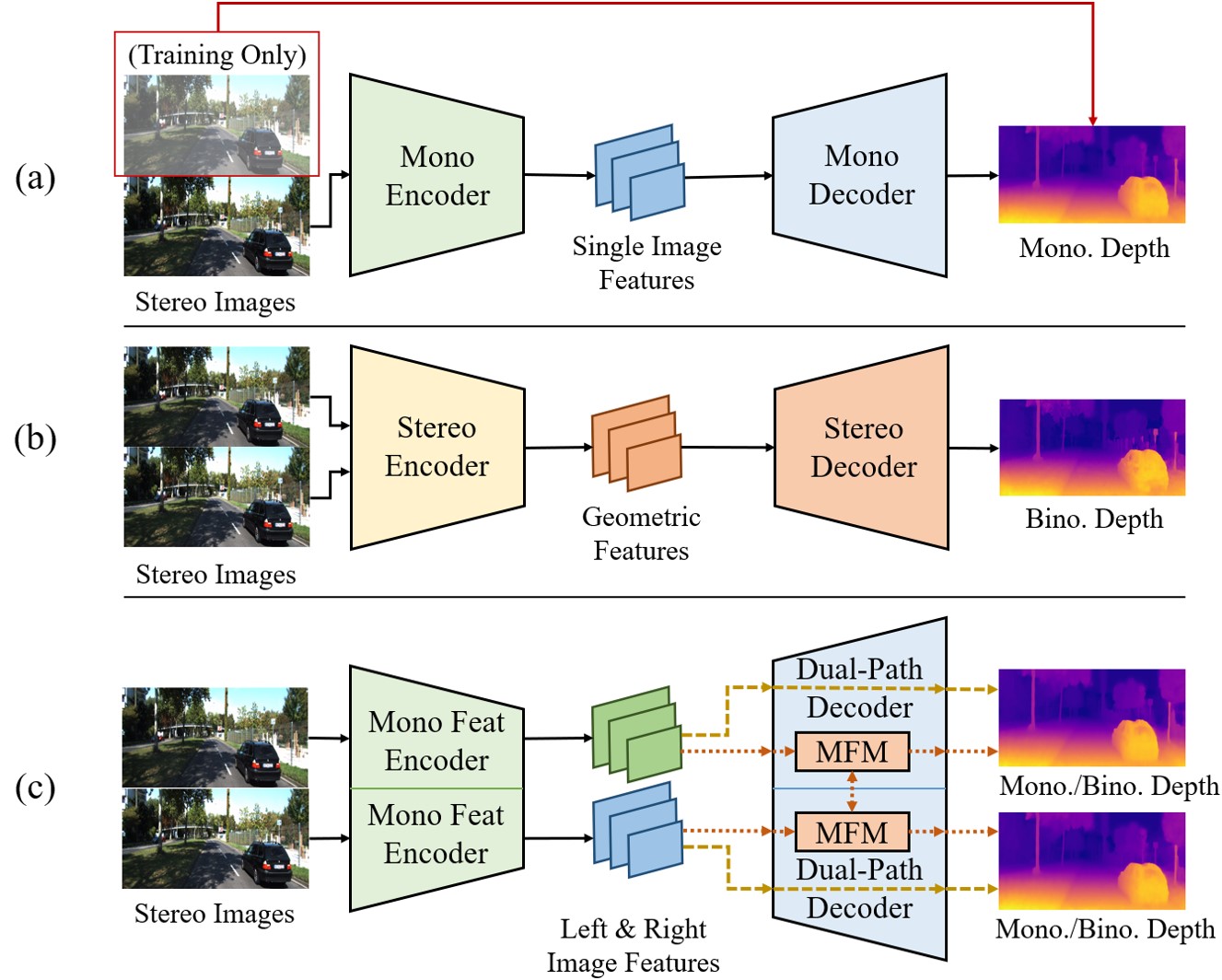}}
  \end{center}
  \caption{Diagrams of three kinds of self-supervised depth estimation models trained with stereo pairs:
  (a) Monocular model is tested with a single image but needs stereo pairs during training.
  (b) Binocular model is trained and tested with stereo pairs, but could not predict depths from a single image; 
  (c) TiO-Depth could be tested with both single images and stereo pairs.}
  \label{fig:method}
\end{figure}

With the development of deep learning techniques, deep-neural-network-based methods have shown their effectiveness for handling both the monocular and binocular depth estimation tasks, which pursue depths from single images and stereo image pairs respectively~\cite{chang2018pyramid, eigen2014depth, fu2018deep, zhang2019ga}.
Since it is time-consuming and labor-intensive to obtain abundant high-quality ground truth scene depths, monocular and binocular self-supervised depth estimation methods, which do not require ground truth depths for training, have attracted increasing attention in recent years~\cite{garg2016unsupervised, godard2019digging, wang2019unos, yang2018segstereo}.

It is noted that the above two tasks are closely related, as shown in~\cref{fig:method}: both the monocular and binocular methods output the same type of results (\ie, depth maps), and some self-supervised monocular methods~\cite{chen2021revealing, godard2017unsupervised, watson2019self} use the same type of training data (\ie, stereo pairs) as the binocular models.
Their main difference is that the monocular task is to predict depths from a single image, while the binocular task is to predict depths from a stereo pair.
Due to this difference, the two tasks have been handled separately by two different kinds of models (\ie, monocular and binocular models) in literature.
Compared with the monocular models that learn depths from single image features, the binocular models focus on learning depths from the geometric features (\eg cost volumes~\cite{yang2018segstereo}) generated with stereo pairs, and consequently, they generally perform better than the monocular models but could not predict depth from a single image.
Moreover, it is found in~\cite{chen2021revealing} that although the whole performances of the monocular models are poorer than the binocular ones, the monocular models still perform better on some special local regions, \eg, the occluded regions around objects which could only be seen at a single view. 
Inspired by this finding, some monocular (or binocular) models employed a separate binocular (or monocular) model to boost their performances in their own task~\cite{aleotti2020reversing, chen2021revealing, choi2021adaptive, facil2017single, long2022two, martins2018fusion, saxena2007depth}. 
All the above issues naturally raise the following problem: \textbf{Is it feasible to explore a general model that could not only compatibly handle the two tasks, but also improve the prediction accuracy?}

Obviously, a general model has the following potential advantages in comparison to the separate models:
\noindent\textbf{(1) Flexibility}: This model could compatibly deal with both the monocular and binocular tasks, and it would be of great benefit to the platforms with a binocular system in the real application, where one camera in the binocular system might be occasionally occluded or even broken down.
\noindent\textbf{(2) High Efficiency}: This model has the potential to perform better than both monocular and binocular models, while the number of its parameters is less than that of two separate models.

Addressing the aforementioned problem and potential advantages of a general depth estimation model, in this paper, we propose a Two-in-One model for both monocular and binocular self-supervised depth estimations, called TiO-Depth.
TiO-Depth employs a monocular model as a sub-network of a Siamese architecture, so that the whole architecture could take stereo images as input.
Considering that the two sub-networks extract image features independently, we design a monocular feature matching module to fuse features from the two sub-networks for binocular prediction.
Then, a multi-stage joint-training strategy is proposed for training TiO-Depth in a self-supervised manner and boosting its accuracy in the two tasks by combining their relative advantages and alleviating their disadvantages.

In sum, our main contributions include:
\begin{itemize}
\item We propose a novel self-supervised depth estimation model called TiO-Depth, which could handle both the monocular and binocular depth estimation tasks.
\item We design a dual-path decoder with the monocular feature matching modules for aggregating the features from either single images or stereo pairs, which may provide new insights into the design of the self-supervised depth estimation network. 
\item We propose a multi-stage joint-training strategy for training TiO-Depth, which is helpful for improving the performances of TiO-Depth in the two tasks.
\end{itemize}

\begin{figure*}[t]
  \begin{center}
   \includegraphics[width=0.9\textwidth]{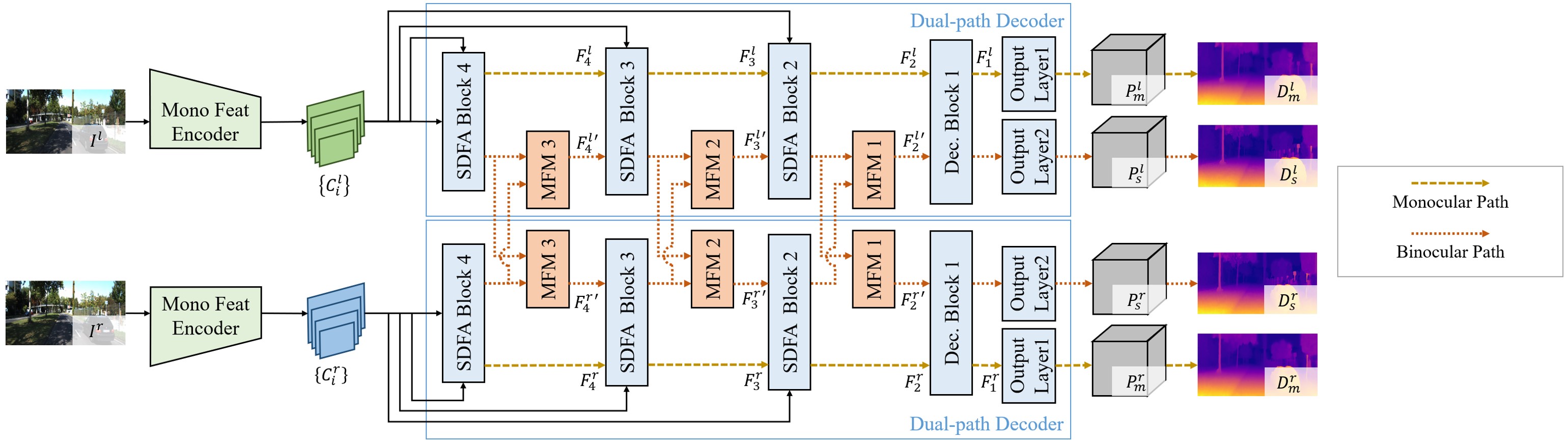}
  \end{center}
   \caption{Architecture of TiO-Depth.
   TiO-Depth employs a Siamese architecture and each sub-network is comprised of a {\bf Mono}cular {\bf Feat}ure {\bf Encoder} and a dual-path decoder.
   The features extracted by the encoder are passed through the decoder via different paths for handling different tasks.
   $\{P_m,P_s\}$ denote the probability volumes predicted by the monocular and binocular paths respectively, while $\{D_m,D_s\}$ are the corresponding depth maps.
   The superscripts `l' and `r' denote the left and right views respectively.
   }
 \label{fig:arc}
\end{figure*}

\section{Related work}
\subsection{Self-supervised monocular depth estimation}
Self-supervised monocular depth estimation methods take multi-view images as training data and learn to estimate the depth from a single input image with the image reconstruction.
The existing methods could be categorized into two groups according to the training data: video training methods and stereo training methods. 

The methods trained with video sequences~\cite{chen2019self,cheng2020s,godard2019digging,guizilini2020semantically,jung2021fine,klingner2020self,petrovai2022exploiting,shu2020feature-metric,yin2018geonet,zhou2017unsupervised,he2022ra} needed to estimate scene depths and camera poses simultaneously.
Zhou \etal~\cite{zhou2017unsupervised} proposed an end-to-end framework which is comprised of two separate networks for predicting depths and camera poses.
Godard \etal~\cite{godard2019digging} designed a per-pixel minimum reprojection loss with an auto-mask and a full-resolution sampling for training the model to learn more accurate depths.
SD-SSMDE~\cite{petrovai2022exploiting} utilized a self-distillation framework where a student network was trained by the absolute depth pseudo labels generated with a teacher network.
Several methods~\cite{cheng2020s,guizilini2020semantically,jung2021fine,klingner2020self} used extra semantic information for improving the performance, and the frameworks explored in~\cite{chen2019self,yin2018geonet} jointly learnt depth, camera pose and optical flow.
Additionally, the multi-frame monocular depth estimation was handled in~\cite{guizilini2022multi,watson2021temporal}, which predicted more accurate depths by taking two frames of a monocular video as input.

The methods trained with stereo image pairs~\cite{bello2021self,chen2021revealing,choi2021adaptive,garg2016unsupervised,godard2017unsupervised,gonzalezbello2020forget, peng2021excavating, pilzer2019refine,tosi2019learning,watson2019self,zhu2020the,zhou2022self,zhou2022learning} generally predicted scene depths by estimating the disparity between the stereo pair.
Godard \etal~\cite{godard2017unsupervised} designed a left-right disparity consistency loss to improve its robustness.
Zhu \etal~\cite{zhu2020the} proposed an edge consistency loss between the depth map and the semantic segmentation map, while a stereo occlusion mask was proposed for alleviating the influence of the occlusion problem during training.
An indirect way of learning depths was proposed in~\cite{bello2021self,gonzalezbello2020forget, gonzalez2021plade}, where the model outputted a probability volume of a set of discrete disparities for depth prediction. 
The self-distillation technique~\cite{gou2021knowledge} was incorporated in~\cite{peng2021excavating,zhou2022self} to boost the performance of the model by using the reliable results predicted by itself.
Considering that the stereo pairs were available at the training stage, Watson \etal~\cite{watson2019self} proposed to utilize the disparities generated with Semi Global Matching~\cite{hirschmuller2005accurate} as the `Depth Hints' to improve the accuracy.
The frameworks that trained a monocular depth estimation network with the pseudo labels selected from the results of a binocular depth estimation network were proposed in~\cite{choi2021adaptive, chen2021revealing}.

\subsection{Self-supervised binocular depth estimation}
Binocular depth estimation (so called as stereo matching) aims to estimate depths by taking stereo image pairs as input~\cite{bleyer2011patchmatch, chang2018pyramid, hirschmuller2005accurate, zhang2019ga}.
Recently, self-supervised binocular depth estimation methods~\cite{zhou2017unsupervised, yang2018segstereo, wang2019unos,liu2020flow2stereo,wang2020parallax,huang2022h, aleotti2020reversing} were proposed for overcoming the limitation of the ground truth. 
Zhou \etal~\cite{zhou2017unsupervised} proposed a framework for learning stereo matching in an iterative manner, which was guided by the left-right check.
UnOS~\cite{wang2019unos} and Flow2Stereo~\cite{liu2020flow2stereo} were proposed for predicting optical flow and binocular depth simultaneously, where the geometrical consistency between the two types of the predicted results was used to improve the accuracy of them.
Wang \etal~\cite{wang2020parallax} proposed a parallax-attention mechanism to learn the stereo correspondence.
H-Net~\cite{huang2022h} was proposed to learn binocular depths with a Siamese network and an epipolar attention mechanism.

\section{Methodology}
In this section, we firstly introduce the architecture of the proposed TiO-Depth, including the details of the dual-path decoder and the Monocular Feature Matching (MFM) module.
Then, we describe the multi-stage joint-training strategy and the loss functions for training TiO-Depth. 
\subsection{Overall architecture}
Since TiO-Depth is to handle both monocular and binocular depth estimation tasks, it should be able to predict depths from both single image features and geometric features, while the binocular and monocular models could only estimate depths from one type of the features respectively.
To this end, TiO-Depth utilizes a Siamese architecture as shown in~\cref{fig:arc}, and each of the two sub-networks is used as a monocular model.
They predict the monocular depth $D_m$ from a single image $I \in \mathbb{R}^{3\times H\times W}$ for avoiding the model learning depths only based on the geometric features, where $\{H, W\}$ denote the height and width of the image.
The parameters of the two sub-networks are shared, and they consist of a monocular feature encoder and a decoder.
For effectively extracting geometric features from available stereo pairs for the binocular task, the dual-path decoder is proposed as the decoder part of the sub-networks, where a binocular path is added to the path for the monocular task (called monocular path).
In the binocular path, the MFM modules are added to learn the geometric features by matching the monocular features extracted by the two sub-networks from a stereo pair and integrate them into the input features.
Accordingly, the full TiO-Depth is used to predict binocular depths $\{D^l_s, D^r_s\}$.

Specifically, a modified Swin-transformer~\cite{liu2021swin} is adopted as the encoder as done in~\cite{zhou2022self}, which extracts 4 image features $\{C_i\}^{4}_{i=1}$ with the resolutions of $\{\frac{H}{2^{i}} \times \frac{W}{2^{i}}\}^{4}_{i=1}$.
We detail the dual-path decoder and the MFM module as following.

\subsection{Dual-path decoder} 
\label{sec:dpd}
As shown in~\cref{fig:arc}, the dual-path decoder is used to gradually aggregate the extracted image features for depth prediction, which consists of three Self-Distilled Feature Aggregation (SDFA) blocks~\cite{zhou2022self}, one decoder block~\cite{godard2019digging}, three monocular feature matching (MFM) modules, and two $3 \times 3$ convolutional layers used as the output layers. 
The features could be passed through different modules via different paths for the monocular and binocular tasks.

For monocular depth estimation, the multi-scale features $\{C_i\}^{4}_{i=1}$ are gradually aggregated by the SDFA blocks and the decoder block, which is defined as the monocular path.
The SDFA block was proposed in~\cite{zhou2022self} for aggregating the features with two resolutions and maintaining the contextual consistency, which takes a low resolution decoder feature $F_{i+1}$ (Specifically, $F_5$ = $C_4$) and a high resolution encoder feature $C_{i-1}$, outputting a new decoder feature with the same shape as $C_{i-1}$.
The decoder block is comprised of two $3 \times 3$ convolutional layers with the ELU activation~\cite{clevert2015fast} and an upsample operation for generating a high resolution feature $F_i$ from the output of the last block.
The output layer is to generate a discrete disparity volume $V \in \mathbb{R}^{N\times H \times W}$ from the last decoder feature $F_1$, where $N$ is the number of the discrete disparity levels.

It is noted that two volumes (defined as the auxiliary volume $V_a$ and the final volume $V_m$) could be generated for monocular depth estimation by using different offset learning branches in SDFA blocks at the training stage, which would be trained with the photometric loss and the distilled loss at different steps respectively. 
More details would be described in~\cref{sec:strategy}.
Accordingly, the branches in SDFA used to generate the two volumes are called auxiliary branch and the final branch.
Since $V_a$ is only used at the training stage, it is not illustrated in~\cref{fig:arc}, and the depth calculated based on $V_m$ is the final monocular result.

For binocular depth estimation, the dual-path decoders in the two sub-networks are utilized for processing left and right image features via the binocular path.
In this path, MFM modules take the decoder features $\{F^{l}_{i}, F^{r}_{i}\}^{4}_{i=2}$ outputted by the SDFA blocks (where the auxiliary branch is used) for generating the corresponding stereo features $\{{F^{l}_{i}}', {F^{r}_{i}}'\}^{4}_{i=2}$ by incorporating the stereo knowledge.
The left and right stereo discrete disparity volumes $\{V^l_s, V^r_s\}$ are obtained by passing the last decoder features $\{F^{l}_{1}, F^{r}_{1}\}$ to another output layer in each decoder.

For obtaining the depth map from the discrete disparity volume $V$, as done in~\cite{bello2020forget,zhou2022self}, a set of discrete disparity levels $\{b_n\}_{n=0}^{N-1}$ is generated with the mirrored exponential disparity discretization by given the maximum and minimum disparities $[b_{\rm min}, b_{\rm max}]$.
Then, a probability volume $ P $ is obtained by normalizing $V$ through a softmax operation along the first (\ie channel) dimension, and a disparity map is calculated by weighted summing of $\{b_n\}_{n=0}^{N-1}$ with the corresponding channels in $P$:
\begin{equation}
   d = \sum_{n=0}^{N-1}P_{n} \odot b_n \quad,
   \label{equ:disp}
\end{equation}
where $P_{n}$ denotes the $n^{\rm th}$ channel of $P$ and `$\odot$' is the element-wise multiplication.
Given the baseline length $B$ of the stereo pair and the horizontal focal length $f_x$ of the camera, the depth map is calculated via $D = \frac{Bf_x}{d}$.

\begin{figure}[t]
  \footnotesize
  \begin{center}
  \centerline{\includegraphics[width=0.8\linewidth]{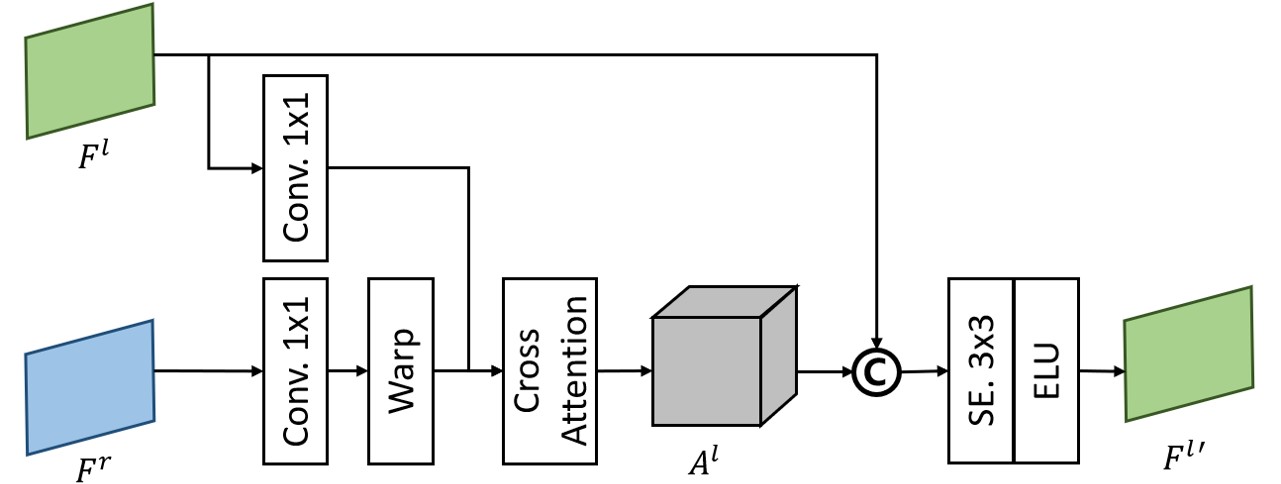}}
  \end{center}
  \caption{Architecture of Monocular Feature Matching (MFM) module.
          `\textcircled{c}' denotes the concatenation operation and `SE.' is the SE convolutional layer~\cite{hu2018squeeze}.}
  \label{fig:module}
\end{figure}

\subsection{Monocular Feature Matching (MFM) module}
Given the features $\{F^l, F^r\}\in \mathbb{R}^{C \times H' \times W'}$ obtained from the two decoders of the two sub-networks, MFM utilizes the cross-attention mechanism~\cite{vaswani2017attention} for generating the cost volume at the left (or right) view and integrates it into the corresponding feature for outputting a stereo feature that has the same shape of input the feature. 
$\{C, H', W'\}$ are the channel, height, and width of the features.
Without loss of generality, as shown in~\cref{fig:module}, for obtaining the stereo feature at the left-view ${F^{l}}'$, MFM firstly applies two $1\times 1$ convolutional layers to generate the left-view query feature $Q^l$ and the right-view key feature $K^r$ from $\{F^l, F^r\}$ respectively.
As done in~\cite{guizilini2022multi}, the left-view cost volume is generated based on the attention scores between $Q^l$ and a set of shifted $K^r$, where each score map $S^l_{n}\in \mathbb{R}^{1 \times H' \times W'}$ is calculated between $Q^l$ and $K^r$ shifted with $b_n'$, which is formulated as:
\begin{equation}
  S^l_{n} = \frac{{\rm sum}(Q^l \odot K_n^r)}{\sqrt{C}} \quad,
\end{equation}
where $K_n^r$ denotes the $K^r$ shifted with $b_n'$, and `${\rm sum}(\cdot)$' is a sum operation along the first dimension.
Then, the cost volume $A^l \in \mathbb{R}^{N \times H' \times W'}$ is obtained by concatenating $S^l_{n}$ generated with all the disparity levels $\{b_n'=\frac{W'}{W}b_n\}^{N-1}_{n=0}$ and normalizing it with a softmax operation along the first dimension:
\begin{equation}
  A^{l} = {\rm softmax}\left([\{S^r_n\}^{N-1}_{n=0}]\right) \quad,
\end{equation}
where `$[\cdot]$' denotes the concatenation operation.
For integrating the stereo knowledge in the cost volume into the decoder feature to obtain the stereo feature ${F^l}'$, $F^l$ and $A^l$ are concatenated and passed through a $3 \times 3$ SE convolutional layer~\cite{hu2018squeeze} with the ELU activation:
\begin{equation}
  {F^l}' = {\rm SE}\left([A^l, F^l]\right) \quad.
\end{equation}

\begin{figure*}[t]
  \begin{center}
   \includegraphics[width=1\textwidth]{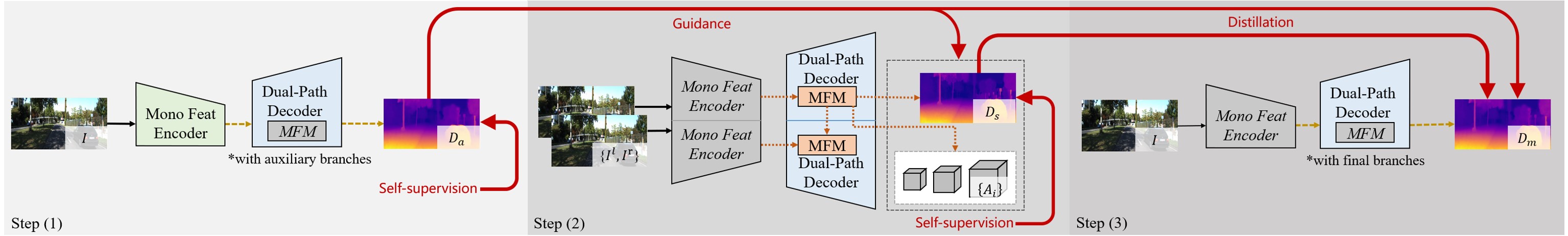}
  \end{center}
   \caption{Multi-stage joint-training strategy.
            There are three steps in each training iteration, where TiO-Depth is trained for different tasks.
            The training at the current step could be benefited from the results generated at the previous steps.
            The modules that do not optimized in each step are denoted by grey and the \emph{italic font}.  
            }
 \label{fig:loss}
\end{figure*}

\subsection{Multi-stage joint-training strategy}
\label{sec:strategy}
TiO-Depth is trained with stereo image pairs in a self-supervised manner.
Considering the motivation of the architecture of TiO-Depth and the different advantages and constraints of the two tasks, we design the multi-stage training strategy as shown in~\cref{fig:loss}.
There are three stages in the strategy, where the training iterations are divided into one, two and three steps respectively.
At the last two stages, the training at the current step could be benefited from the results generated at the previous steps.
We detail the three steps as following.

\textbf{Step (1)}. TiO-Depth is trained for learning monocular depth estimation under monocular constraints at this step.
The discrete depth constraint~\cite{zhou2022learning,bello2020forget} is used to generate a left-view reconstructed image $\hat{I}^l_a$ with the right-view auxiliary volume $V^r_a$ (generated with the auxiliary branches in SDFAs as mentioned in~\cref{sec:dpd}) and the right-view real image $I^r$.
As done in~\cite{bello2020forget,zhou2022self}, the monocular loss $L_{M}$ for training TiO-Depth contains a reconstruction loss $L_{rec1}$ for reflecting the difference between $\hat{I}^l_a$ and $I^l$, and an edge-aware smoothness loss $L_{smo1}$:
\begin{equation}
  L_{M} = L_{rec1} + \lambda_1 L_{smo1},
\end{equation}
where $\lambda_1$ is a preset weight parameters.
All the parameters in TiO-Depth except MFMs are optimized at this step.

\textbf{Step (2)}. TiO-Depth is trained for learning binocular depth estimation under binocular constraints and some monocular results obtained at step (1).
The continuous depth constraint~\cite{zhou2022learning, chen2021revealing} is used to reconstruct a left-view image $\tilde{I}^l_s$ by taking the right-view image $I^r$ and the predicted left-view depth map $D^l_s$ as the input.
Then, a stereo loss is adopted to train the network, which consists of the following terms:

The stereo reconstruction loss term $L_{rec2}$ is formulated as a weighted sum of the $L_1$ loss and the structural similarity (SSIM) loss~\cite{wang2004image} as done in~\cite{chen2021revealing,godard2019digging}.
Considering the relative advantage of the monocular results on the occluded regions, the occluded pixels in $I^l$ are replaced by the corresponding pixels in a monocular reconstructed image $\tilde{I}^l_a$ calculated with the auxiliary monocular depth map $D^l_a$:
\begin{equation}
   L_{rec2} = \alpha \left \| \tilde{I}^l_s -  {I^l}' \right \|_1
     +(1 - \alpha) {\rm SSIM}{(\tilde{I}^l_s, {I^l}')} \quad,
\end{equation}
\begin{equation}
   {I^l}' = M^l_{occ} \odot I^l + (1 - M^l_{occ}) \odot \tilde{I}^l_a \quad,
\end{equation}
where $\alpha$ is a balance parameter and `$\| \cdot \|_1$' denotes the $L_1$ norm.
$M^l_{occ}$ is an occlusion mask generated with the auxiliary monocular disparity $d^l_a$ as done in~\cite{zhu2020the}, where the values are zeros in the occluded regions, and ones otherwise.

The cost volume loss term $L_{cos}$ is adopted to guide the cost volumes $\{A^{l}_i\}^3_{i=1}$ generated in MFMs through the auxiliary monocular probability volume $P^{l}_{a}$, which is formulated as:
\begin{equation}
   L_{cos} = \sum^{3}_{i=1}{\frac{1}{\Omega_i}\sum_{\left \| A^{l}_i(x)-P^{l}_{a}\left <x \right > \right \|_1 > t_1}\left \| A^{l}_i(x)-P^{l}_{a} \left <x\right > \right \|_1},
\end{equation}
where $\Omega_i$ denotes the number of the valid coordinates $x$ in $A_i$, and $t_1$ is a predefined threshold.
`$\left< \cdot \right>$' denotes the bilinear sampling operation for getting the element at the corresponding coordinate of $x$ in a different resolution volume.

The disparity guidance loss term $L_{gui}$ leverages both the gradient information and the edge region values in the auxiliary monocular disparity map $d^{l}_a$ for improving the quality of the binocular result:
\begin{align}
   L_{gui}&= \left \| \partial_x d^{l}_a - \partial_x d^{l}_s \right \|_1
   + \left \| \partial_y d^{l}_a - \partial_y d^{l}_s \right \|_1 \nonumber \\
   &+ M^l_{out} \odot \left \|  d^{l}_a - d^{l}_s \right \|_1
   \quad,
\end{align}
where `$\partial_x$', `$\partial_y$' are the differential operators in the horizontal and vertical directions respectively, $M^l_{out}$ denotes a binary mask~\cite{mahjourian2018unsupervised} where the pixels whose reprojected coordinates are out of the image are ones, and zeros otherwise.
Accordingly, the stereo loss is formulated as:
\begin{equation}
  L_{S} = L_{rec2} + \lambda_2 L_{smo2} + \lambda_3 L_{cos} + \lambda_4 L_{gui} \quad,
\end{equation}
where $\{\lambda_2, \lambda_3, \lambda_4\}$ are preset weight parameters, and $L_{smo2}$ is the edge-aware smoothness loss~\cite{godard2019digging}.
At this step, only the parameters in the dual-path decoder are optimized.

\begin{table*}[]
  \begin{center}
  \footnotesize
  \renewcommand\tabcolsep{3pt}
  \begin{tabular}{|lccc|cccc|ccc|}
  \hline
  Method &
    PP. &
    Sup. &
    Resolution &
    Abs. Rel. $\downarrow$ &
    Sq. Rel. $\downarrow$ &
    RMSE $\downarrow$ &
    logRMSE $\downarrow$ &
    A1 $\uparrow$ &
    A2 $\uparrow$ &
    A3 $\uparrow$ \\\hline
  R-MSFM6~\cite{zhou2021r} &
    &
   M &
   320$\times$1024 &
   0.108 &
   0.748 &
   4.470 &
   0.185 &
   0.889 &
   0.963 &
   0.982 \\
  PackNet~\cite{guizilini20203d} &
     &
    M &
    384$\times$1280 &
    0.107 &
    0.802 &
    4.538 &
    0.186 &
    0.889 &
    0.962 &
    0.981 \\
  SGDepth~\cite{klingner2020self} &
        &
    M(Se.) &
    384$\times$1280 &
    0.107 &
    0.768 &
    4.468 &
    0.186 &
    0.891 &
    0.963 &
    0.982 \\
  SD-SSMDE~\cite{petrovai2022exploiting} &
     &
    M &
    320$\times$1024 &
    0.098 &
    0.674 &
    4.187 &
    0.170 &
    0.902 &
    0.968 &
    0.985 \\\hline
  monoResMatch~\cite{tosi2019learning} &
    \checkmark &
    S(SGM) &
    384$\times$1280 &
    0.111 &
    0.867 &
    4.714 &
    0.199 &
    0.864 &
    0.954 &
    0.979 \\
  Monodepth2~\cite{godard2019digging} &
    \checkmark &
    S &
    320$\times$1024 &
    0.105 &
    0.822 &
    4.692 &
    0.199 &
    0.876 &
    0.954 &
    0.977 \\
  DepthHints~\cite{watson2019self} &
    \checkmark &
    S(SGM) &
    320$\times$1024 &
    0.096 &
    0.710 &
    4.393 &
    0.185 &
    0.890 &
    0.962 &
    0.981 \\
  SingleNet~\cite{chen2021revealing} &
    \checkmark &
    S(S.T.) &
    320$\times$1024 &
    0.094 &
    0.681 &
    4.392 &
    0.185 &
    0.892 &
    0.962 &
    0.981 \\
  FAL-Net~\cite{gonzalezbello2020forget} &
    \checkmark &
    S &
    384$\times$1280 &
    0.093 &
    0.564 &
    3.973 &
    0.174 &
    0.898 &
    0.967 &
    \textbf{0.985} \\
  Edge-of-depth~\cite{zhu2020the} &
    \checkmark &
    S(SGM, Se.) &
    320$\times$1024 &
    0.091 &
    0.646 &
    4.244 &
    0.177 &
    0.898 &
    0.966 &
    0.983 \\
  PLADE-Net~\cite{gonzalez2021plade} &
    \checkmark &
    S &
    384$\times$1280 &
    0.089 &
    0.590 &
    4.008 &
    0.172 &
    0.900 &
    0.967 &
    \textbf{0.985} \\
  EPCDepth~\cite{peng2021excavating} &
    \checkmark &
    S(SGM) &
    320$\times$1024 &
    0.091 &
    0.646 &
    4.207 &
    0.176 &
    0.901 &
    0.966 &
    0.983 \\
  OCFD-Net~\cite{zhou2022learning} &
    \checkmark &
    S & 
    384$\times$1280 &
    0.090 & 
    0.563 & 
    4.005 & 
    0.172 & 
    0.903 & 
    0.967 & 
    \underline{0.984} \\
  SDFA-Net~\cite{zhou2022self} &
  \checkmark  &
    S &
    384$\times$1280 &
    0.089 &
    \underline{0.531} &
    \textbf{3.864} &
    \underline{0.168} &
    0.907 &
    \underline{0.969} &
    \textbf{0.985} \\
  \emph{TiO-Depth} &  & S & 384$\times$1280 &
  \underline{0.085} & 0.544 & \underline{3.919} & 0.169 & \underline{0.911} & \underline{0.969} & \textbf{0.985} \\
  \emph{TiO-Depth} & \checkmark & S & 384$\times$1280 &
  \textbf{0.083} & \textbf{0.521} & \textbf{3.864} & \textbf{0.167} & \textbf{0.912} & \textbf{0.970} & \textbf{0.985} \\\hline
  DepthFormer (2F.)~\cite{guizilini2022multi} &  & M & 192$\times$640  & 0.090 & 0.661 & 4.149 & 0.175 & 0.905 & 0.967 & 0.984 \\
  ManyDepth (2F.)~\cite{watson2021temporal}   &  & M & 320$\times$1024 & 0.087 & 0.685 & 4.142 & 0.167 & 0.920 & 0.968 & 0.983  \\ 
  H-Net (Bino.)~\cite{huang2022h} &
     &
    S &
    192$\times$640 &
    0.076 &
    0.607 &
    4.025 &
    0.166 &
    0.918 &
    0.966 &
    0.982 \\
  \emph{TiO-Depth (Bino.)} & & S & 384$\times$1280 &
  \textbf{0.063} & \textbf{0.523} & \textbf{3.611} & \textbf{0.153} & \textbf{0.943} & \textbf{0.972} & \textbf{0.985} \\\hline
  \end{tabular}
\end{center}
\caption{Quantitative comparison on the KITTI Eigen test set.
          $\downarrow/\uparrow$ denotes that lower / higher is better.
          The best and the second best results are in \textbf{bold} and \underline{underlined} under each metric.
          The methods marked with `2F.' predict depths by taking 2 frames from a monocular video as input, while the methods with `Bino.' predict depths by taking stereo pairs as input.
          `PP.' means using the post-processing step.
          The methods marked with `Se.', `SGM', and `S.T.' are trained with the semantic segmentation label, the depth generated with SGM~\cite{hirschmuller2005accurate}, and the depth predicted by a binocular teacher network respectively.
          }
  \label{tab:main}
\end{table*}

\textbf{Step (3)}. 
TiO-Depth is trained in a distilled manner by utilizing the results obtained at step (1)\&(2) as the teacher for further improving monocular prediction.
A distilled loss $L_{dis}$ is used to constrain the final monocular probability volume $P^{l}_m$ (generated with the final branches in SDFAs) with the stereo probability volume $P^{l}_s$ and the auxiliary monocular probability volume $P^{l}_a$. 
Considering the relative advantages of the monocular and stereo results, a hybrid probability volume $P^{l}_h$ is generated by fusing them weighted by a half-object-edge map $M^l_{hoe}$:
\begin{equation}
  P^{l}_h = (1 - M^l_{hoe}) \odot P^{l}_s +M^l_{hoe} \odot P^{l}_a \quad.
\end{equation}
$M^l_{hoe}$ is a grayscale map for indicating the flat areas and the areas on one side of the
object, where the binocular results are more accurate experimentally:
\begin{equation}
  M^l_{hoe} = M^l_{occ'} \odot {\rm min}(\frac{{\rm maxpool}(\|k * D^l_s\|_1)}{t_2}, 1) \quad,
\end{equation}
where `${\rm maxpool}(\cdot)$' denotes a $3\times 3$ max pooling layer with stride 1, `$*$' denotes the convolutional operation, $k$ is a $3\times 3$ Laplacian kernel, and $t_2$ is a predefined threshold.
$M^l_{occ'}$ is an opposite occlusion mask obtained by treating the left-view disparity map as the right-view one during calculating the occlusion mask.
KL divergence is employed to reflect the similarity between the final monocular probability volume $P^{l}_m$ and $P^{l}_h$, which is formulated as:
\begin{equation}
  L_{dis} = {\rm KL}(P^{l}_h||P^{l}_m) \quad.
\end{equation}
Only the parameters in the SDFA blocks, the decoder block and the output layer are optimized at this step.
Please see the supplemental material for more details about the training strategy and losses.

\section{Experiments}
In this section, we train TiO-Depth on the KITTI dataset~\cite{geiger2012we}, and the evaluations are conducted on the KITTI, Cityscapes~\cite{cordts2016cityscapes}, and DDAD~\cite{guizilini20203d} datasets.
For monocular depth estimation, the Eigen split~\cite{eigen2014depth} of KITTI is utilized, which consists of a training set with 22600 stereo pairs and a test set with 697 images.
For binocular depth estimation, a training set with 28968 stereo pairs collected from KITTI is used for training as done in~\cite{chen2021revealing,liu2019unsupervised,wang2019unos}, while the training set of the KITTI 2015 stereo benchmark~\cite{menze2015joint} is used for the evaluation, which consists of 200 image pairs. 
For exploring the generation ability of TiO-Depth, Cityscapes and DDAD are used for conducting an additional evaluation.
Please see the supplemental material for more details about the datasets and metrics.

\begin{figure*}[]
  \begin{center}
  \footnotesize
  \includegraphics[width=0.97\textwidth]{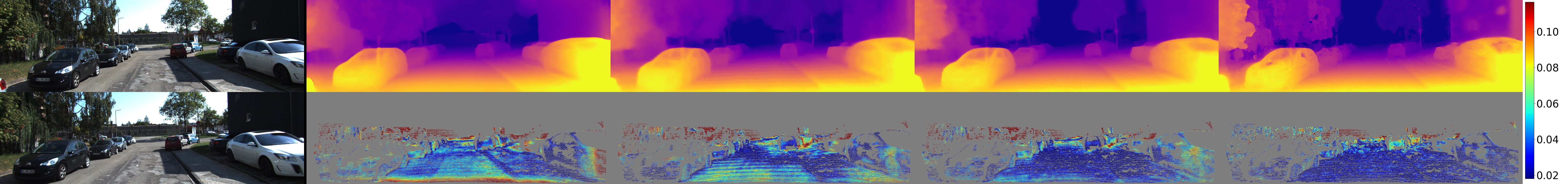}
  \includegraphics[width=0.97\textwidth]{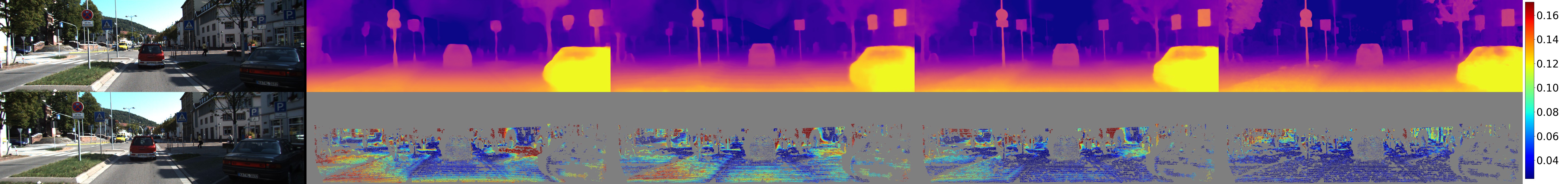}
  \includegraphics[width=0.97\textwidth]{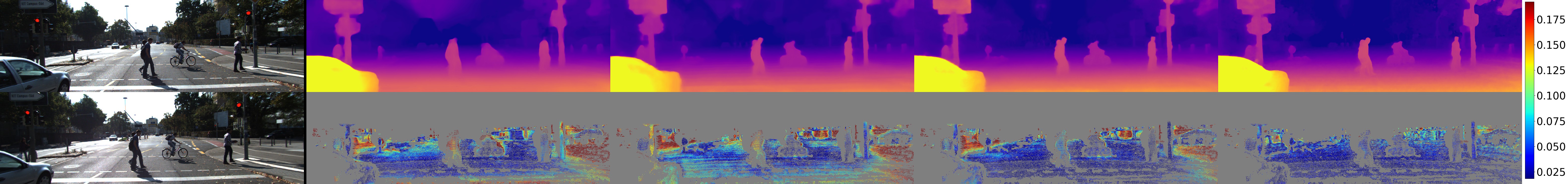}
  \leftline{\qquad \qquad Input Images
            \qquad \qquad \qquad \enspace EPCDepth~\cite{peng2021excavating}
            \qquad \qquad \quad SDFA-Net~\cite{zhou2022self}
            \qquad \qquad  TiO-Depth (Mono.)
            \qquad \qquad TiO-Depth (Bino.)}
  \end{center}
  \caption{Visualization results of EPCDepth~\cite{peng2021excavating}, SDFA-Net~\cite{zhou2022self} and our TiO-Depth on KITTI.
  The input stereo pairs are shown in the first column, where the left-view images are used for monocular depth estimation.
  The predicted depth maps with the corresponding `Abs. Rel.' error maps calculated on an improved Eigen test set~\cite{uhrig2017sparsity} are shown in the following columns.
  For the error maps, red indicates larger error, and blue indicates smaller error as shown in the color bars.}
\label{fig:comp}
\end{figure*}

\begin{table*}[]
  \begin{center}
  \footnotesize
  \renewcommand\tabcolsep{1.5pt}
  \begin{tabular}{|lcc|cccc|ccc|cc|}
  \hline
  Method &
  Sup. &
  Resolution &
  Abs. Rel. $\downarrow$ &
  Sq. Rel. $\downarrow$ &
  RMSE $\downarrow$ &
  logRMSE $\downarrow$ &
  A1 $\uparrow$ &
  A2 $\uparrow$ &
  A3 $\uparrow$ &
  EPE-all$\downarrow$ &
  D1-all$\downarrow$ \\\hline
  MonoDepth~\cite{godard2017unsupervised}  &S & 256$\times$512   & 0.068             & 0.835             & 4.392             & 0.146             & 0.942             & 0.978             & 0.989             & -                 & 9.194 \\
  UnOS (Stereo-only)~\cite{wang2019unos}   &S & 256$\times$832   & 0.060             & 0.833             & 4.187             & 0.135             & 0.955             & 0.981             & 0.990             & -                 & 7.073 \\
  UnOS (Full)~\cite{wang2019unos}          &MS & 256$\times$832   & \underline{0.049} & 0.515             & 3.404             & 0.121             & 0.965 & 0.984             & \underline{0.992} & -                 & \textbf{5.943} \\
  Liu \etal~\cite{liu2019unsupervised}    &S & 256$\times$832   & 0.051             & 0.532             & 3.780             & 0.126             & 0.957             & 0.982             & 0.991             & 1.520             & 9.570 \\
  Flow2Stereo~\cite{liu2020flow2stereo}    &MS & 384$\times$1280  & -                 & -                 & -                 & -                 & -                 & -                 & -                 & \underline{1.340}    & \underline{6.130} \\
  StereoNet~\cite{chen2021revealing}       &S & 320$\times$1024  & 0.052             & 0.558             & 3.733             & 0.123             & 0.961             & 0.984             & \underline{0.992} & -                 & -     \\
  StereoNet-D~\cite{chen2021revealing}     &S* & 320$\times$1024  & \textbf{0.048}    & \underline{0.482} & \underline{3.393} & \underline{0.105}    & \textbf{0.969}    &\textbf{ 0.989}    & \textbf{0.994}    & -                 & -     \\
  \emph{TiO-Depth}                     &S & 384$\times$1280  & 0.050             & \textbf{0.434}    & \textbf{3.239}    & \textbf{0.104} & \underline{0.967}             & \underline{0.987} &\textbf{0.994}             & \textbf{1.282} & 6.647    \\\hline
  SingleNet (Mono.)~\cite{chen2021revealing}    &S(S.T.)  & 320$\times$1024 & 0.083          & 0.688          & 4.464          & 0.154          & 0.904          & 0.972          & 0.990          & -              & -               \\
  \emph{TiO-Depth (Mono.)} &S & 384$\times$1280 & 0.075          & 0.458          & 3.717          & 0.130          & \textbf{0.925}          & 0.979          & 0.992          & 2.203          & 17.860          \\
  \emph{TiO-Depth (Mono.)+PP.} &S & 384$\times$1280 & \textbf{0.073} & \textbf{0.439} & \textbf{3.680} & \textbf{0.128} & \textbf{0.925} & \textbf{0.980} & \textbf{0.993} & \textbf{2.158} & \textbf{17.570}  \\\hline
  \end{tabular}
\end{center}
 \caption{Quantitative comparison on KITTI 2015 training set.
           The methods marked with `Mono.' predict depths by taking single image as input, while other methods predict depths with stereo pairs.
           `S*' denotes the method is jointly trained with a separate monocular model.}
  \label{tab:stereo}
\end{table*}

\subsection{Implementation details}
TiO-Depth is implemented with the PyTorch~\cite{paszke2019pytorch} framework.
The tiny size modified Swin-transformer~\cite{liu2021swin,zhou2022self} used as the monocular feature encoder is pretrained on the ImageNet dataset~\cite{russakovsky2015imagenet}.
We set the minimum and the maximum disparities to $b_{\min}=2,b_{\max}=300$ for the discrete disparity volume, and the number of the discrete disparity levels is set to $N=49$.
The weight parameters for the loss function are set to $\lambda_1=0.0008,\lambda_2=0.008, \lambda_3=0.01$, and $\lambda_4=0.01$, while we set $\alpha=0.15$ $t_1 = 1$, and $t_2 = 0.13$.
The Adam optimizers~\cite{kingma2014adam} with $\beta_1=0.5$ and $\beta_2=0.999$ are used to train TiO-Depth for 50 epochs.
The learning rate is firstly set to $10^{-4}$, and is downgraded by half at the 20, 30, 40, 45 epochs.
At both the training and testing stages, the images are resized into the resolution of $384 \times 1280 $, while we assume that the intrinsics of all the images are identical. 
The on-the-fly data augmentations are performed in training, including random resizing (from 0.67 to 1.5) and cropping (256$\times$832), random horizontal flipping, and random color augmentation.

\subsection{Comparative evaluation}
For monocular depth estimation, we firstly evaluate TiO-Depth on the KITTI Eigen test set~\cite{eigen2014depth} in comparison to 4 methods trained with monocular video sequences (M) and 10 methods trained with stereo image pairs (S). 
The corresponding results by all the referred methods are cited from their original papers and reported in~\cref{tab:main}.

It can be seen that TiO-Depth with a post-processing as done in~\cite{zhou2022self} outperforms all the comparative methods in most cases, including the methods trained with the depth pseudo labels generated by additional algorithms or networks (SGM, S.T.).
Since \emph{the same} TiO-Depth model could handle the binocular task by using the binocular path, we give its performance in binocular depth estimation (`Bino.') in comparison with 3 methods.
As seen from~\cref{tab:main}, TiO-Depth gets the top performance among all the comparative multi-frame (2F.) and binocular methods.
Several visualization results of TiO-Depth as well as two comparative methods: EPCDepth~\cite{peng2021excavating} and SDFA-Net~\cite{zhou2022self} are given in~\cref{fig:comp}.
As shown in the figure, the depth maps predicted by TiO-Depth are more accurate and contain more delicate geometric details, while the performance of TiO-Depth is further improved by taking the stereo pairs as input.
These results demonstrate that the TiO-Depth could predict accurate depths by taking both monocular and binocular inputs.

For binocular depth estimation, we evaluate TiO-Depth on the KITTI 2015 training set~\cite{menze2015joint} in comparison to 5 self-supervised binocular depth estimation methods.
It is noted that all of the comparative methods could not handle the monocular task.
As seen from the corresponding results shown in~\cref{tab:stereo}, TiO-Depth outperforms all the methods trained with stereo pairs (S) or stereo videos (MS) in most cases, and it achieves comparable performance with StereoNet-D~\cite{chen2021revealing} benefited from an additional monocular depth estimation model, while the performance of TiO-Depth is boosted by itself.
The monocular depth estimation results of \emph{the same} TiO-Depth model are also given in~\cref{tab:stereo}, which show that it effectively handling the monocular task at the same time, further indicating the effectiveness of TiO-Depth as a two-in-one model.

Furthermore, we train TiO-Depth on KITTI~\cite{geiger2012we} and evaluate it on DDAD~\cite{guizilini20203d} and Cityscapes~\cite{cordts2016cityscapes} for testing its cross-dataset generalization ability. 
The corresponding results of TiO-Depth and 6 comparative methods are reported in~\cref{tab:gen}.
As shown in the table, TiO-Depth not only performs best in comparison to the methods evaluated in a cross-dataset manner, but also achieves a competitive performance with the methods trained and tested on the same dataset.
When the stereo pairs are available, TiO-Depth could predict more accurate binocular depths by taking the image pairs.
These results demonstrate the generalization ability of TiO-Depth on the unseen dataset.
Please see the supplemental material for the additional exponential results.

\begin{table}[]
  \begin{center}
  \footnotesize
  \renewcommand\tabcolsep{1.4pt}
  \begin{tabular}{|lcc|ccc|c|}
    \hline
        Method & train & test & Abs. Rel. $\downarrow$ & Sq. Rel. $\downarrow$ & RMSE $\downarrow$ & A1 $\uparrow$ \\ \hline
        PackNet~\cite{guizilini20203d} & D & D & 0.173 & 7.164 & 14.363 & 0.835 \\ 
        ManyDepth (2F.)~\cite{watson2021temporal} & D & D & 0.146 & 3.258 & 14.098 & 0.822 \\
        DepthFormer (2F.)~\cite{guizilini2022multi} & D & D & \textbf{0.135} & 2.953 & \textbf{12.477} & \textbf{0.836} \\ 
        \emph{TiO-Depth} & K & D & 0.144 & \textbf{2.664} & 14.273 & 0.808 \\ \hline\hline
        MonoDepth2~\cite{godard2019digging} & C & C & 0.129 & 1.569 & 6.876 & 0.849 \\
        Li \etal~\cite{li2021unsupervised} & C & C & 0.119 & 1.290 & 6.980 & 0.846 \\ 
        ManyDepth (2F.)~\cite{watson2021temporal} & C & C & \textbf{0.114} & 1.193 & 6.223 & \textbf{0.875} \\ 
        SD-SSMDE~\cite{petrovai2022exploiting} & C & C & \textbf{0.114} & \textbf{1.017} & \textbf{5.949} & 0.870 \\ \hline
        MonoDepth2~\cite{godard2019digging} & K & C & 0.153 & 1.785 & 8.590 & 0.774 \\ 
        SD-SSMDE~\cite{petrovai2022exploiting} & K & C & 0.143 & 1.635 & 8.441 & 0.789 \\ 
        \emph{TiO-Depth} & K & C & \textbf{0.120} & \textbf{1.176} & \textbf{7.157} & \textbf{0.850} \\ \hline
        \emph{TiO-Depth (Bino.)} & K & C & 0.066 & 0.423 & 4.070 & 0.961 \\ \hline
    \end{tabular}
  \end{center}
    \caption{Quantitative comparison on DDAD and Cityscapes.
  `C', `K', and `D' denote the methods are trained or tested on the Cityscapes, KITTI and DDAD datasets respectively.}
  \label{tab:gen}
\end{table}

\begin{table}[]
  \begin{center}
  \footnotesize
  \renewcommand\tabcolsep{2.3pt}
  \begin{tabular}{|l|cc|c|cc|}
    \hline
    Methods & Abs. Rel. $\downarrow$ & Sq. Rel. $\downarrow$ & A1 $\uparrow$ & EPE $\downarrow$ & D1 $\downarrow$ \\ \hline
        w. Cat module (321) & 0.069 & 0.505 & 0.947 & 2.074 & 15.952 \\
        w. Attn module (321)& 0.053 & 0.439 & 0.965 & 1.377 & 7.421 \\ \hline
        w. MFM (1) & 0.054 & \textbf{0.423} & 0.960 & 1.483 & 8.784 \\ 
        w. MFM (21) & 0.052 & 0.445 & 0.965 & 1.305 & 7.077 \\
        TIO-Depth & \textbf{0.051} & 0.429 & \textbf{0.966} & \textbf{1.281} & \textbf{6.684} \\ \hline
        w/o. $L_{gui}$& 0.053 & 0.506 & \textbf{0.966} & 1.292 & 6.984 \\
        w/o. $L_{gui}$, $L_{cos}$ & 0.053 & 0.522 & 0.965 & 1.326 & 6.755 \\
        w/o. $L_{gui}$, $L_{cos}$, $M_{occ}$ & 0.054 & 0.565 & 0.963 & 1.345 & 7.159 \\
        \hline
  \end{tabular}
\end{center}
  \caption{Binocular depth estimation results on KITTI 2015 training set in the ablation study.
  The numbers in the name of methods mean the indexes of the used modules as shown in~\cref{fig:arc}.
  All the results are evaluated after training 30 epochs.}
  \label{tab:abla-st2-1}
\end{table}

\begin{table}[]
  \begin{center}
  \footnotesize
  \renewcommand\tabcolsep{4.5pt}
  \begin{tabular}{|ccc|ccc|c|}
    \hline
    Steps & $L_{dis}$ & FB. & Abs. Rel. $\downarrow$ & Sq. Rel. $\downarrow$ & RMSE $\downarrow$ & A1 $\uparrow$ \\ \hline
    1 & - & - & 0.088 & 0.556 & 4.093 & 0.904 \\
    1+2 & - & - & 0.088 & 0.557 & 4.067 & 0.906 \\
    1+2+3 &$P^l_s$& \checkmark& 0.086 & 0.590 & 4.021 & \textbf{0.911} \\
    1+2+3 &$P^l_h$& \checkmark& \textbf{0.085} & \textbf{0.544} & \textbf{3.919} & \textbf{0.911} \\
    1+2+3 &$P^l_h$& - & 0.098 & 0.695 & 4.367 & 0.892 \\ \hline
        
    \end{tabular}
  \end{center}
    \caption{Monocular depth estimation results on the KITTI Eigen test set in the ablation study.
             'FB.' denotes using the final branches.}
  \label{tab:abla-st}
\end{table}

\subsection{Ablation studies}
This subsection verifies the effectiveness of each key element in TiO-Depth by conducting ablation studies on the KITTI dataset~\cite{geiger2012we}. 

\textbf{Dual-path decoder.}
We firstly replace the proposed Monocular Feature Matching (MFM) modules with the concatenation-based modules (Cat module) and the cross-attention-based modules without the SE layer (Attn modules), respectively.
The corresponding results are shown in the first part of \cref{tab:abla-st2-1}, which show that TiO-Depth (with MFM (321)) performs best compared to the models with other modules.
Then, the impact of the number of MFMs is shown in the second part of~\cref{tab:abla-st2-1}.
It can be seen that the binocular performances are gradually improved by using more MFMs in most cases.
The monocular depth estimation results of TiO-Depth with/without the `final branch (FB.)' in the SDFA modules are shown in the last two rows of~\cref{tab:abla-st}, where the performance of TiO-Depth with the final branches is much better than that of the model without these branches.
We notice that the switchable branches are important for TiO-Depth to improve the monocular results, but the SDFA block is not a necessary choice.
Please see the supplemental material for more experimental results and discussions.
Considering that the three MFMs only contain 1.7M parameters in total, these results indicate the effectiveness of the dual-path decoder with MFMs in the two tasks.

\textbf{Multi-stage joint-training strategy.}
We firstly analyze the impact of each term in the stereo loss $L_S$ in binocular depth estimation by sequentially taking out the disparity guidance loss term $L_{gui}$, the cost volume loss term $L_{cos}$ and the occlusion mask $M_{occ}$ used in $L_{rec2}$.
The corresponding results in the third part of~\cref{tab:abla-st2-1} show that the performances of the model are dropped by removing the loss terms and the mask.
Then we train TiO-Depth with different numbers of step(s) and pseudo labels to validate the effectiveness of the training strategy in monocular depth estimation in~\cref{tab:abla-st}. 
As shown in the table, the monocular performance could not be improved by just training TiO-Depth for learning the two tasks without distillation (\ie, with `1+2' steps), but it is improved in most cases by training with three steps.
Compared with using the stereo probability volume $P^l_s$, the accuracy of the monocular results could be consistently improved by using the hybrid probability volume $P^l_h$ in the distilled loss $L_{dis}$.
These results demonstrate that our training strategy is helpful for TiO-Depth to learn more accurate monocular and binocular depths.

\section{Conclusion}
In this paper, we propose TiO-Depth, a two-in-one depth prediction model for both the monocular and binocular self-supervised depth estimation tasks, while a multi-stage joint-training strategy is explored for training.
The full TiO-Depth is used to predict depths from stereo pairs, while the partial TiO-Depth by closing the duplicate parts could predict depths from single images. 
The experimental results in monocular and binocular depth estimations not only prove the effectiveness of TiO-Depth but also indicate the feasibility of bridging the gap between the two tasks. 

\noindent
{\bf Acknowledgements.}
This work was supported by the Strategic Priority Research Program of the Chinese Academy of Sciences (Grant No. XDA27040811), the National Natural Science Foundation of China (Grant Nos. 61991423, U1805264), the Beijing Municipal Science and Technology Project (Grant No. Z211100011021004).

{\small
\bibliographystyle{ieee_fullname}
\bibliography{egbib}

\begin{thebibliography}{10}\itemsep=-1pt

\bibitem{aleotti2020reversing}
Filippo Aleotti, Fabio Tosi, Li Zhang, Matteo Poggi, and Stefano Mattoccia.
\newblock Reversing the cycle: self-supervised deep stereo through enhanced
  monocular distillation.
\newblock In {\em European Conference on Computer Vision}, pages 614--632.
  Springer, 2020.

\bibitem{bello2020forget}
Juan Luis~Gonzalez Bello and Munchurl Kim.
\newblock Forget about the lidar: Self-supervised depth estimators with med
  probability volumes.
\newblock {\em Advances in Neural Information Processing Systems}, 33, 2020.

\bibitem{bello2021self}
Juan Luis~Gonzalez Bello and Munchurl Kim.
\newblock Self-supervised deep monocular depth estimation with ambiguity
  boosting.
\newblock {\em IEEE TPAMI}, 2021.

\bibitem{bleyer2011patchmatch}
Michael Bleyer, Christoph Rhemann, and Carsten Rother.
\newblock Patchmatch stereo-stereo matching with slanted support windows.
\newblock In {\em Bmvc}, volume~11, pages 1--11, 2011.

\bibitem{chang2018pyramid}
Jia-Ren Chang and Yong-Sheng Chen.
\newblock Pyramid stereo matching network.
\newblock In {\em CVPR}, pages 5410--5418, 2018.

\bibitem{chen2019self}
Yuhua Chen, Cordelia Schmid, and Cristian Sminchisescu.
\newblock Self-supervised learning with geometric constraints in monocular
  video: Connecting flow, depth, and camera.
\newblock In {\em ICCV}, pages 7063--7072, 2019.

\bibitem{chen2021revealing}
Zhi Chen, Xiaoqing Ye, Wei Yang, Zhenbo Xu, Xiao Tan, Zhikang Zou, Errui Ding,
  Xinming Zhang, and Liusheng Huang.
\newblock Revealing the reciprocal relations between self-supervised stereo and
  monocular depth estimation.
\newblock In {\em ICCV}, pages 15529--15538, 2021.

\bibitem{cheng2020s}
Bin Cheng, Inderjot~Singh Saggu, Raunak Shah, Gaurav Bansal, and Dinesh
  Bharadia.
\newblock S3 net: Semantic-aware self-supervised depth estimation with
  monocular videos and synthetic data.
\newblock In {\em ECCV}, pages 52--69, 2020.

\bibitem{choi2021adaptive}
Hyesong Choi, Hunsang Lee, Sunkyung Kim, Sunok Kim, Seungryong Kim, Kwanghoon
  Sohn, and Dongbo Min.
\newblock Adaptive confidence thresholding for monocular depth estimation.
\newblock In {\em ICCV}, pages 12808--12818, 2021.

\bibitem{clevert2015fast}
Djork-Arn{\'e} Clevert, Thomas Unterthiner, and Sepp Hochreiter.
\newblock Fast and accurate deep network learning by exponential linear units
  (elus).
\newblock {\em arXiv preprint arXiv:1511.07289}, 2015.

\bibitem{cordts2016cityscapes}
Marius Cordts, Mohamed Omran, Sebastian Ramos, Timo Rehfeld, Markus Enzweiler,
  Rodrigo Benenson, Uwe Franke, Stefan Roth, and Bernt Schiele.
\newblock The cityscapes dataset for semantic urban scene understanding.
\newblock In {\em Proceedings of the IEEE conference on computer vision and
  pattern recognition}, pages 3213--3223, 2016.

\bibitem{dai2017deformable}
Jifeng Dai, Haozhi Qi, Yuwen Xiong, Yi Li, Guodong Zhang, Han Hu, and Yichen
  Wei.
\newblock Deformable convolutional networks.
\newblock In {\em Proceedings of the IEEE international conference on computer
  vision}, pages 764--773, 2017.

\bibitem{eigen2015predicting}
David Eigen and Rob Fergus.
\newblock Predicting depth, surface normals and semantic labels with a common
  multi-scale convolutional architecture.
\newblock In {\em Proceedings of the IEEE international conference on computer
  vision}, pages 2650--2658, 2015.

\bibitem{eigen2014depth}
David Eigen, Christian Puhrsch, and Rob Fergus.
\newblock Depth map prediction from a single image using a multi-scale deep
  network.
\newblock {\em Advances in neural information processing systems}, 27, 2014.

\bibitem{facil2017single}
Jos{\'e}~M F{\'a}cil, Alejo Concha, Luis Montesano, and Javier Civera.
\newblock Single-view and multi-view depth fusion.
\newblock {\em IEEE Robotics and Automation Letters}, 2(4):1994--2001, 2017.

\bibitem{fu2018deep}
Huan Fu, Mingming Gong, Chaohui Wang, Kayhan Batmanghelich, and Dacheng Tao.
\newblock Deep ordinal regression network for monocular depth estimation.
\newblock In {\em CVPR}, pages 2002--2011, 2018.

\bibitem{garg2016unsupervised}
Ravi Garg, Vijay~Kumar BG, Gustavo Carneiro, and Ian Reid.
\newblock Unsupervised cnn for single view depth estimation: Geometry to the
  rescue.
\newblock In {\em ECCV}, pages 740--756, 2016.

\bibitem{geiger2012we}
Andreas Geiger, Philip Lenz, and Raquel Urtasun.
\newblock Are we ready for autonomous driving? the kitti vision benchmark
  suite.
\newblock In {\em CVPR}, pages 3354--3361, 2012.

\bibitem{godard2017unsupervised}
Cl{\'e}ment Godard, Oisin Mac~Aodha, and Gabriel~J Brostow.
\newblock Unsupervised monocular depth estimation with left-right consistency.
\newblock In {\em CVPR}, pages 270--279, 2017.

\bibitem{godard2019digging}
Cl{\'e}ment Godard, Oisin Mac~Aodha, Michael Firman, and Gabriel~J Brostow.
\newblock Digging into self-supervised monocular depth estimation.
\newblock In {\em ICCV}, pages 3828--3838, 2019.

\bibitem{gonzalezbello2020forget}
Juan~Luis GonzalezBello and Munchurl Kim.
\newblock Forget about the lidar: Self-supervised depth estimators with med
  probability volumes.
\newblock {\em Advances in Neural Information Processing Systems},
  33:12626--12637, 2020.

\bibitem{gonzalez2021plade}
Juan~Luis GonzalezBello and Munchurl Kim.
\newblock Plade-net: Towards pixel-level accuracy for self-supervised
  single-view depth estimation with neural positional encoding and distilled
  matting loss.
\newblock In {\em CVPR}, pages 6851--6860, 2021.

\bibitem{gordon2019depth}
Ariel Gordon, Hanhan Li, Rico Jonschkowski, and Anelia Angelova.
\newblock Depth from videos in the wild: Unsupervised monocular depth learning
  from unknown cameras.
\newblock In {\em Proceedings of the IEEE/CVF International Conference on
  Computer Vision}, pages 8977--8986, 2019.

\bibitem{gou2021knowledge}
Jianping Gou, Baosheng Yu, Stephen~J Maybank, and Dacheng Tao.
\newblock Knowledge distillation: A survey.
\newblock {\em IJCV}, 129(6):1789--1819, 2021.

\bibitem{guizilini20203d}
Vitor Guizilini, Rares Ambrus, Sudeep Pillai, Allan Raventos, and Adrien
  Gaidon.
\newblock 3d packing for self-supervised monocular depth estimation.
\newblock In {\em CVPR}, pages 2485--2494, 2020.

\bibitem{guizilini2022multi}
Vitor Guizilini, Rareș Ambruș, Dian Chen, Sergey Zakharov, and Adrien Gaidon.
\newblock Multi-frame self-supervised depth with transformers.
\newblock In {\em CVPR}, pages 160--170, 2022.

\bibitem{guizilini2020semantically}
Vitor Guizilini, Rui Hou, Jie Li, Rares Ambrus, and Adrien Gaidon.
\newblock Semantically-guided representation learning for self-supervised
  monocular depth.
\newblock In {\em International Conference on Learning Representations (ICLR)},
  2020.

\bibitem{he2022ra}
Mu He, Le Hui, Yikai Bian, Jian Ren, Jin Xie, and Jian Yang.
\newblock Ra-depth: Resolution adaptive self-supervised monocular depth
  estimation.
\newblock In {\em Computer Vision--ECCV 2022: 17th European Conference, Tel
  Aviv, Israel, October 23--27, 2022, Proceedings, Part XXVII}, pages 565--581.
  Springer, 2022.

\bibitem{hirschmuller2005accurate}
Heiko Hirschmuller.
\newblock Accurate and efficient stereo processing by semi-global matching and
  mutual information.
\newblock In {\em CVPR}, volume~2, pages 807--814. IEEE, 2005.

\bibitem{hu2018squeeze}
Jie Hu, Li Shen, and Gang Sun.
\newblock Squeeze-and-excitation networks.
\newblock In {\em Proceedings of the IEEE conference on computer vision and
  pattern recognition}, pages 7132--7141, 2018.

\bibitem{huang2022h}
Baoru Huang, Jian-Qing Zheng, Stamatia Giannarou, and Daniel~S Elson.
\newblock H-net: Unsupervised attention-based stereo depth estimation
  leveraging epipolar geometry.
\newblock In {\em CVPR}, pages 4460--4467, 2022.

\bibitem{johnson2016perceptual}
Justin Johnson, Alexandre Alahi, and Li Fei-Fei.
\newblock Perceptual losses for real-time style transfer and super-resolution.
\newblock In {\em ECCV}, pages 694--711, 2016.

\bibitem{jung2021fine}
Hyunyoung Jung, Eunhyeok Park, and Sungjoo Yoo.
\newblock Fine-grained semantics-aware representation enhancement for
  self-supervised monocular depth estimation.
\newblock In {\em ICCV}, pages 12642--12652, 2021.

\bibitem{kingma2014adam}
Diederik~P Kingma and Jimmy Ba.
\newblock Adam: A method for stochastic optimization.
\newblock {\em arXiv preprint arXiv:1412.6980}, 2014.

\bibitem{klingner2020self}
Marvin Klingner, Jan-Aike Term{\"o}hlen, Jonas Mikolajczyk, and Tim
  Fingscheidt.
\newblock Self-supervised monocular depth estimation: Solving the dynamic
  object problem by semantic guidance.
\newblock In {\em ECCV}, pages 582--600, 2020.

\bibitem{li2021unsupervised}
Hanhan Li, Ariel Gordon, Hang Zhao, Vincent Casser, and Anelia Angelova.
\newblock Unsupervised monocular depth learning in dynamic scenes.
\newblock In {\em Conference on Robot Learning}, pages 1908--1917. PMLR, 2021.

\bibitem{liu2019unsupervised}
Liang Liu, Guangyao Zhai, Wenlong Ye, and Yong Liu.
\newblock Unsupervised learning of scene flow estimation fusing with local
  rigidity.
\newblock In {\em IJCAI}, pages 876--882, 2019.

\bibitem{liu2020flow2stereo}
Pengpeng Liu, Irwin King, Michael~R Lyu, and Jia Xu.
\newblock Flow2stereo: Effective self-supervised learning of optical flow and
  stereo matching.
\newblock In {\em CVPR}, pages 6648--6657, 2020.

\bibitem{liu2021swin}
Ze Liu, Yutong Lin, Yue Cao, Han Hu, Yixuan Wei, Zheng Zhang, Stephen Lin, and
  Baining Guo.
\newblock Swin transformer: Hierarchical vision transformer using shifted
  windows.
\newblock In {\em ICCV}, pages 10012--10022, 2021.

\bibitem{long2022two}
Yangqi Long, Huimin Yu, and Biyang Liu.
\newblock Two-stream based multi-stage hybrid decoder for self-supervised
  multi-frame monocular depth.
\newblock {\em IEEE Robotics and Automation Letters}, 7(4):12291--12298, 2022.

\bibitem{mahjourian2018unsupervised}
Reza Mahjourian, Martin Wicke, and Anelia Angelova.
\newblock Unsupervised learning of depth and ego-motion from monocular video
  using 3d geometric constraints.
\newblock In {\em Proceedings of the IEEE/CVF Conference on Computer Vision and
  Pattern Recognition (CVPR)}, pages 5667--5675, 2018.

\bibitem{martins2018fusion}
Diogo Martins, Kevin Van~Hecke, and Guido De~Croon.
\newblock Fusion of stereo and still monocular depth estimates in a
  self-supervised learning context.
\newblock In {\em 2018 IEEE International Conference on Robotics and Automation
  (ICRA)}, pages 849--856. IEEE, 2018.

\bibitem{menze2015joint}
Moritz Menze, Christian Heipke, and Andreas Geiger.
\newblock Joint 3d estimation of vehicles and scene flow.
\newblock In {\em ISPRS Workshop on Image Sequence Analysis (ISA)}, 2015.

\bibitem{paszke2019pytorch}
Adam Paszke, Sam Gross, Francisco Massa, Adam Lerer, James Bradbury, Gregory
  Chanan, Trevor Killeen, Zeming Lin, Natalia Gimelshein, Luca Antiga, et~al.
\newblock Pytorch: An imperative style, high-performance deep learning library.
\newblock {\em Advances in neural information processing systems},
  32:8026--8037, 2019.

\bibitem{peng2021excavating}
Rui Peng, Ronggang Wang, Yawen Lai, Luyang Tang, and Yangang Cai.
\newblock Excavating the potential capacity of self-supervised monocular depth
  estimation.
\newblock In {\em ICCV}, pages 15560--15569, 2021.

\bibitem{petrovai2022exploiting}
Andra Petrovai and Sergiu Nedevschi.
\newblock Exploiting pseudo labels in a self-supervised learning framework for
  improved monocular depth estimation.
\newblock In {\em CVPR}, pages 1578--1588, 2022.

\bibitem{pilzer2019refine}
Andrea Pilzer, Stephane Lathuiliere, Nicu Sebe, and Elisa Ricci.
\newblock Refine and distill: Exploiting cycle-inconsistency and knowledge
  distillation for unsupervised monocular depth estimation.
\newblock In {\em CVPR}, pages 9768--9777, 2019.

\bibitem{russakovsky2015imagenet}
Olga Russakovsky, Jia Deng, Hao Su, Jonathan Krause, Sanjeev Satheesh, Sean Ma,
  Zhiheng Huang, Andrej Karpathy, Aditya Khosla, and Michael Bernstein.
\newblock Imagenet large scale visual recognition challenge.
\newblock {\em International Journal of Computer Vision}, 11(3):211--252, 2015.

\bibitem{saxena2007depth}
Ashutosh Saxena, Jamie Schulte, Andrew~Y Ng, et~al.
\newblock Depth estimation using monocular and stereo cues.
\newblock In {\em IJCAI}, volume~7, pages 2197--2203, 2007.

\bibitem{shu2020feature-metric}
Chang Shu, Kun Yu, Zhixiang Duan, and Kuiyuan Yang.
\newblock Feature-metric loss for self-supervised learning of depth and
  egomotion.
\newblock In {\em ECCV}, pages 572--588, 2020.

\bibitem{simonyan2014very}
Karen Simonyan and Andrew Zisserman.
\newblock Very deep convolutional networks for large-scale image recognition.
\newblock {\em arXiv preprint arXiv:1409.1556}, 2014.

\bibitem{tosi2019learning}
Fabio Tosi, Filippo Aleotti, Matteo Poggi, and Stefano Mattoccia.
\newblock Learning monocular depth estimation infusing traditional stereo
  knowledge.
\newblock In {\em CVPR}, pages 9799--9809, 2019.

\bibitem{uhrig2017sparsity}
Jonas Uhrig, Nick Schneider, Lukas Schneider, Uwe Franke, Thomas Brox, and
  Andreas Geiger.
\newblock Sparsity invariant cnns.
\newblock In {\em 2017 international conference on 3D Vision (3DV)}, pages
  11--20, 2017.

\bibitem{vaswani2017attention}
Ashish Vaswani, Noam Shazeer, Niki Parmar, Jakob Uszkoreit, Llion Jones,
  Aidan~N Gomez, {\L}ukasz Kaiser, and Illia Polosukhin.
\newblock Attention is all you need.
\newblock {\em Advances in neural information processing systems}, 30, 2017.

\bibitem{wang2020parallax}
Longguang Wang, Yulan Guo, Yingqian Wang, Zhengfa Liang, Zaiping Lin, Jungang
  Yang, and Wei An.
\newblock Parallax attention for unsupervised stereo correspondence learning.
\newblock {\em IEEE TPAMI}, 2020.

\bibitem{wang2019unos}
Yang Wang, Peng Wang, Zhenheng Yang, Chenxu Luo, Yi Yang, and Wei Xu.
\newblock Unos: Unified unsupervised optical-flow and stereo-depth estimation
  by watching videos.
\newblock In {\em CVPR}, pages 8071--8081, 2019.

\bibitem{wang2004image}
Zhou Wang, Alan~C Bovik, Hamid~R Sheikh, and Eero~P Simoncelli.
\newblock Image quality assessment: from error visibility to structural
  similarity.
\newblock {\em IEEE TIP}, 13(4):600--612, 2004.

\bibitem{watson2019self}
Jamie Watson, Michael Firman, Gabriel~J Brostow, and Daniyar Turmukhambetov.
\newblock Self-supervised monocular depth kints.
\newblock In {\em ICCV}, pages 2162--2171, 2019.

\bibitem{watson2021temporal}
Jamie Watson, Oisin Mac~Aodha, Victor Prisacariu, Gabriel Brostow, and Michael
  Firman.
\newblock The temporal opportunist: Self-supervised multi-frame monocular
  depth.
\newblock In {\em Proceedings of the IEEE/CVF Conference on Computer Vision and
  Pattern Recognition}, pages 1164--1174, 2021.

\bibitem{yang2018segstereo}
Guorun Yang, Hengshuang Zhao, Jianping Shi, Zhidong Deng, and Jiaya Jia.
\newblock Segstereo: Exploiting semantic information for disparity estimation.
\newblock In {\em ECCV}, pages 636--651, 2018.

\bibitem{yin2018geonet}
Zhichao Yin and Jianping Shi.
\newblock Geonet: Unsupervised learning of dense depth, optical flow and camera
  pose.
\newblock In {\em CVPR}, pages 1983--1992, 2018.

\bibitem{zhang2019ga}
Feihu Zhang, Victor Prisacariu, Ruigang Yang, and Philip~HS Torr.
\newblock Ga-net: Guided aggregation net for end-to-end stereo matching.
\newblock In {\em CVPR}, pages 185--194, 2019.

\bibitem{zhou2017unsupervised}
Chao Zhou, Hong Zhang, Xiaoyong Shen, and Jiaya Jia.
\newblock Unsupervised learning of stereo matching.
\newblock In {\em ICCV}, pages 1567--1575, 2017.

\bibitem{zhou2022learning}
Zhengming Zhou and Qiulei Dong.
\newblock Learning occlusion-aware coarse-to-fine depth map for self-supervised
  monocular depth estimation.
\newblock In {\em Proceedings of the 30th ACM International Conference on
  Multimedia}, pages 6386–--6395, 2022.

\bibitem{zhou2022self}
Zhengming Zhou and Qiulei Dong.
\newblock Self-distilled feature aggregation for self-supervised monocular
  depth estimation.
\newblock In {\em ECCV}, pages 709--726. Springer, 2022.

\bibitem{zhou2021r}
Zhongkai Zhou, Xinnan Fan, Pengfei Shi, and Yuanxue Xin.
\newblock R-msfm: Recurrent multi-scale feature modulation for monocular depth
  estimating.
\newblock In {\em ICCV}, pages 12777--12786, 2021.

\bibitem{zhu2020the}
Shengjie Zhu, Garrick Brazil, and Xiaoming Liu.
\newblock The edge of depth: Explicit constraints between segmentation and
  depth.
\newblock In {\em CVPR}, pages 13116--13125, 2020.

\bibitem{zhu2019deformable}
Xizhou Zhu, Han Hu, Stephen Lin, and Jifeng Dai.
\newblock Deformable convnets v2: More deformable, better results.
\newblock In {\em Proceedings of the IEEE/CVF conference on computer vision and
  pattern recognition}, pages 9308--9316, 2019.

\end{thebibliography}
}
\clearpage

\appendix

{\Large\textbf{Supplemental Material}}

\section{Multi-stage joint-training strategy}
\subsection{Image reconstruction}
As mentioned in Sec. 3.4 of the main paper, the discrete depth constraint~\cite{bello2020forget,gonzalez2021plade,bello2021self,zhou2022self} is used for helping TiO-Depth learn monocular depth estimation at step (1), which assumes that the depth of each pixel is inversely proportional to a weighted sum of a set of discrete disparities determined by the visual consistency between the input training stereo images~\cite{zhou2022learning}. 
A left-view reconstructed image $\hat{I}^l_a \in \mathbb{R}^{3\times H \times W}$ is obtained with the right-view real image $I^r \in \mathbb{R}^{3\times H \times W}$ and the predicted right-view auxiliary volume $V^r_a \in \mathbb{R}^{N\times H \times W}$ under the discrete depth constraint, where $N$ is the number of the discrete disparity levels and $\{H, W\}$ are the height and width of the image.
Specifically, a left-view auxiliary volume $\hat{V}^l_a\in \mathbb{R}^{N\times H \times W}$ is firstly generated by shifting the $n^{\rm th}$ channel of $V^r_a$ with the corresponding disparity value $b_n$ generated with  the mirrored exponential disparity discretization~\cite{bello2020forget}.
Then, $\hat{V}^l_a$ is passed thought a softmax operation along the first dimension to obtain the corresponding probability volume $\hat{P}^{l}_{a}$.
Accordingly, the left-view reconstructed image $\hat{I}^l_a$ is obtained by calculating a weighted sum of the shifted $N$ versions of the right image $I^r$ with $\hat{P}^{l}_{a}$:
\begin{equation}
    \hat{I}^l = \sum_{n=0}^{N-1}{\hat{P}^{l}_{an} \odot I^r_{n}}\quad,
\end{equation}
where $\hat{P}^{l}_{an} \in \mathbb{R}^{1\times H \times W}$ is the $n^{\rm th}$ channel of $\hat{P}^{l}_{a}$, `$\odot$' denotes the element-wise multiplication, and $I^r_{n}$ is the left-view image shifted with $b_n$.

The continuous depth constraint~\cite{godard2017unsupervised,godard2019digging,chen2021revealing} is used for helping TiO-Depth learn binocular depth estimation at step (2), which assumes that the depth of each pixel is a continuous variable determined by the visual consistency between the input training stereo images~\cite{zhou2022learning}.
A left-view image $\tilde{I}^l_s$ is obtained with the right-view real image $I^r$ and the predicted left-view depth map $D^l_s \in \mathbb{R}^{1\times H \times W}$ under the continuous depth constraint.
Specifically, for an arbitrary pixel coordinate $p \in \mathbb{R}^2 $ in the left-view image, its corresponding coordinate $p'$ in the right image could be calculated with $D^l_s$:
\begin{equation}
    p'=p-\left[\frac{Bf_x}{D^l_s(p)},0\right]^\top
    \quad,
\end{equation}
where $B$ is the baseline length of the stereo pair and $f_x$ is the horizontal focal length of the camera.
Accordingly, the reconstructed left-view image $\tilde{I}^l_s$ is obtained by assigning the RGB value of the right image pixel $p'$ to the pixel $p$ of $\tilde{I}^l_s$.

\subsection{Monocular loss}
The monocular loss used in step (1) contains a monocular reconstruction loss $L_{rec1}$ and an edge-aware smoothness loss $L_{smo1}$.
Specifically, $L_{rec1}$ consists a $L_1$ loss term and a perceptual loss \cite{johnson2016perceptual} term for measuring the similarity between the left-view reconstructed image $\hat{I}^l_a$ and the left-view real image $I^l$ as done in~\cite{bello2020forget, zhou2022self}:
\begin{equation}
   L_{rec1} = \left \|\hat{I}^l_a - I^l \right \|_1 + \beta \sum_{i=1,2,3} \left \|\phi_i (\hat{I}^l_a) - \phi_i (I^l) \right \|_2
   \quad,
\end{equation}
where `$\| \cdot \|_1$' and `$\| \cdot \|_2$' denote the $L_1$ and $L_2$ norms, $\phi_i(\cdot)$ represents the output of $i^{\rm th}$ pooling layer of a pretrained VGG19~\cite{simonyan2014very}, and $\beta=0.01$ is a balance parameter.
The edge-aware smoothness loss $L_{smo1}$ is employed for constraining the continuity of the auxiliary disparity map $d^r_a$ as done in~\cite{godard2017unsupervised,zhou2022self,bello2020forget,chen2021revealing}:
\begin{equation}
  L_{smo1}= \left \| \partial_x d^r_a \right \|_1 e^{-\gamma \left \| \partial_x I^r \right \|_1}
  + \left \| \partial_y d^r_a \right \|_1 e^{-\gamma \left \| \partial_y I^r \right \|_1}\
  \quad,
  \label{equ:smo}
\end{equation}
where `$\partial_x$', `$\partial_y$' are the differential operators in the horizontal and vertical directions respectively, and $\gamma=2$ is a parameter for adjusting the degree of edge preservation.

\begin{table*}[]
  \begin{center}
  \footnotesize
  \renewcommand\tabcolsep{1.5pt}
  \begin{tabular}{|lccc|cccc|ccc|}
  \hline
  Method &
    PP. &
    Sup. &
    Resolution &
    Abs Rel $\downarrow$ &
    Sq Rel $\downarrow$ &
    RMSE $\downarrow$ &
    logRMSE $\downarrow$ &
    A1 $\uparrow$ &
    A2 $\uparrow$ &
    A3 $\uparrow$ \\\hline
  DepthHints~\cite{watson2019self} &
    \checkmark &
    S(SGM) &
    320$\times$1024 &
    0.074 &
    0.364 &
    3.202 &
    0.114 &
    0.936 &
    0.989 &
    0.997 \\
  FAL-Net~\cite{gonzalezbello2020forget} &
    \checkmark &
    S &
    384$\times$1280 &
    0.071 &
    0.281 &
    2.912 &
    0.108 &
    0.943 &
    0.991 &
    \underline{0.998} \\
  PLADE-Net~\cite{gonzalez2021plade} &
    \checkmark &
    S &
    384$\times$1280 &
    \underline{0.066} &
    0.272 &
    2.918 &
    0.104 &
    0.945 &
    0.992 &
    \underline{0.998} \\
  OCFD-Net~\cite{zhou2022learning} &
    \checkmark &
    S & 
    384$\times$1280 &
    0.069 & 
    0.262 & 
    2.785 & 
    0.103 & 
    0.951 & 
    \underline{0.993}& 
    \underline{0.998} \\
  SDFA-Net~\cite{zhou2022self} &
   \checkmark &
    S &
    384$\times$1280 &
    0.074 &
    \underline{0.228} &
    \textbf{2.547} &
    0.101 &
    0.956 &
    \textbf{0.995} &
    \textbf{0.999} \\
  \emph{TiO-Depth} &  & S & 384$\times$1280 &
  \underline{0.066} & 0.229 & 2.597 & \underline{0.096} & \underline{0.961} & \textbf{0.995} & \textbf{0.999} \\
  \emph{TiO-Depth} & \checkmark & S & 384$\times$1280 &
  \textbf{0.065} & \textbf{0.218} & \underline{2.558} & \textbf{0.094} & \textbf{0.962} & \textbf{0.995} & \textbf{0.999} \\\hline
  DepthFormer (2F.)~\cite{guizilini2022multi} &  & M & 320$\times$1024  & 0.055 & 0.265 & 2.723 & 0.092 & 0.959 & 0.992 & 0.998 \\
  ManyDepth (2F.)~\cite{watson2021temporal}   &  & M & 352$\times$1216 & 0.055 & 0.305 & 2.945 & 0.094 & 0.963 & 0.992 & 0.997  \\ 
  \emph{TiO-Depth (Bino.)} &  & S & 384$\times$1280 &
  \textbf{0.033} & \textbf{0.078} & \textbf{1.583} & \textbf{0.050} & \textbf{0.996} & \textbf{0.999} & \textbf{1.000} \\\hline
  \end{tabular}
\end{center}
\caption{Quantitative comparison on the improved KITTI Eigen test set.
        $\downarrow/\uparrow$ denotes that lower / higher is better.
        The best and the second best results are in \textbf{bold} and \underline{underlined} under each metric.
        The methods marked with `2F.' predict depths by taking 2 frames from a monocular video as input, while the methods with `Bino.' predict depths by taking stereo pairs as input.
        `PP.' means using the post-processing step.
        The methods marked with `SGM' are trained with the the depth generated with SGM~\cite{hirschmuller2005accurate}.
        }
  \label{tab:main2}
\end{table*}

\begin{table*}[]
\begin{center}
\footnotesize
\renewcommand\tabcolsep{1.5pt}
\begin{tabular}{|lcc|cccc|ccc|}
  \hline
  Method                & train & test & Abs. Rel. $\downarrow$ & Sq. Rel. $\downarrow$ & RMSE $\downarrow$ & logRMSE $\downarrow$ & A1 $\uparrow$ & A2 $\uparrow$ & A3 $\uparrow$ \\\hline
  PackNet~\cite{guizilini20203d}               & D     & D    & 0.173                  & 7.164                 & 14.363            & 0.249                & 0.835    & -              & -              \\
ManyDepth (2F.)~\cite{watson2021temporal}   & D     & D    & 0.146                  & 3.258                 & 14.098      & -                    & 0.822          & -              & -              \\
DepthFormer (2F.)~\cite{guizilini2022multi} & D     & D    & \textbf{0.135}         & 2.953           & \textbf{12.477}   & -                    & \textbf{0.836} & -              & -              \\
\emph{TiO-Depth}             & K     & D    & 0.144            & \textbf{2.664}        & 14.273            & \textbf{0.242}       & 0.808          & 0.933          & 0.970        \\\hline\hline
  MonoDepth2~\cite{godard2019digging}            & C     & C    & 0.129                  & 1.569                 & 6.876             & 0.187                & 0.849          & 0.957          & 0.983          \\
Li \etal~\cite{li2021unsupervised}                    & C     & C    & 0.119      & 1.290                 & 6.980             & 0.190                & 0.846          & 0.952         & 0.982          \\
ManyDepth (2F.)~\cite{watson2021temporal}   & C     & C    & \textbf{0.114}         & 1.193     & 6.223 & 0.170     & \textbf{0.875} & \textbf{0.967} & 0.989 \\
SD-SSMDE~\cite{petrovai2022exploiting}              & C     & C    & \textbf{0.114}         & \textbf{1.017}        & \textbf{5.949}    & \textbf{0.169}       & 0.870    & \textbf{0.967} & \textbf{0.990}          \\\hline
MonoDepth2~\cite{godard2019digging}             & K     & C    & 0.153                  & 1.785                 & 8.590             & 0.234                & 0.774          & 0.926          & 0.976          \\
SD-SSMDE~\cite{petrovai2022exploiting}              & K     & C    & 0.143                  & 1.635                 & 8.441             & 0.221                & 0.789          & 0.931          & 0.980          \\
\emph{TiO-Depth}             & K     & C    & \textbf{0.120}      & \textbf{1.176}     & \textbf{7.157}             & \textbf{0.187}                & \textbf{0.850}          & \textbf{0.958}          & \textbf{0.987}    \\\hline
\emph{TiO-Depth (Bino.)}     & K     & C    & 0.066                  & 0.423                 & 4.070             & 0.106                & 0.961          & 0.992          & 0.997          \\\hline
  \end{tabular}
\end{center}
\caption{Quantitative comparison on DDAD~\cite{guizilini20203d} and Cityscapes~\cite{cordts2016cityscapes} (Tab. 3 in the main paper).
`C', `K', and `D' denote the methods are trained or tested on the Cityscapes, KITTI and DDAD datasets respectively.}
\label{tab:gen2}
\end{table*}

\subsection{Details of the training}
Since the predicted depth results are not reliable at the early training epochs, which lack the ability to effectively guide the following steps, the second and third steps are enabled after $E_1=20$ and $E_2=30$ training epochs respectively.
Thus, the multi-stage joint-training strategy contains three stages, where the training iterations are divided into one, two and three steps respectively as mentioned in Sec. 3.4 of the main paper.
Considering that the second and the third steps are enabled after $E_1$ and $E_2$ epochs respectively and different parameters are optimized at these steps, we use three Adam optimizers~\cite{kingma2014adam} at the three steps for training.
The learning rate of each optimizer is set to $10^{-4}$ when the corresponding training step is firstly enabled, and which is downgraded by half as described in Sec. 4.1 of the main paper.
Since there are several parameters are trained only at one step (\eg, the parameters in the monocular feature matching modules), while other parameters are trained at multiple steps (\eg, the parameters in the decoder block), we multiply the learning rates of the parameters that have optimized at the previous steps by 0.1.

\begin{figure*}[]
  \begin{center}
  \footnotesize
  \includegraphics[width=0.97\textwidth]{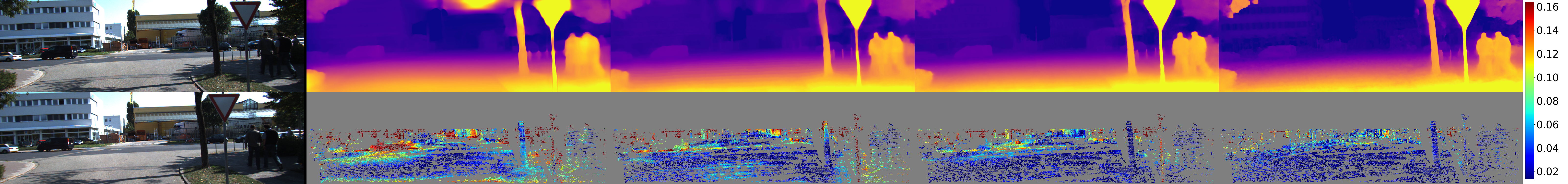}
  \includegraphics[width=0.97\textwidth]{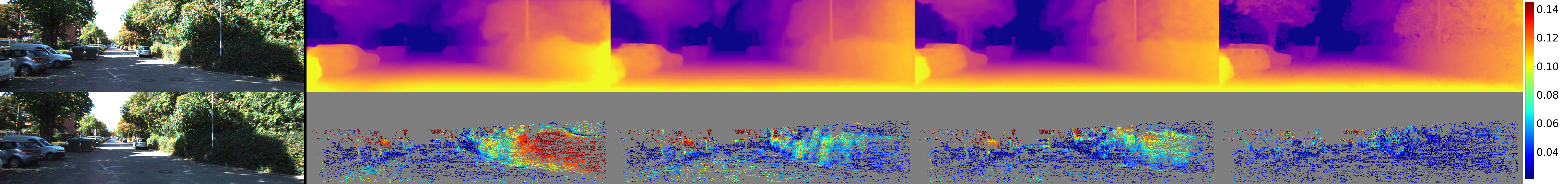}
  \includegraphics[width=0.97\textwidth]{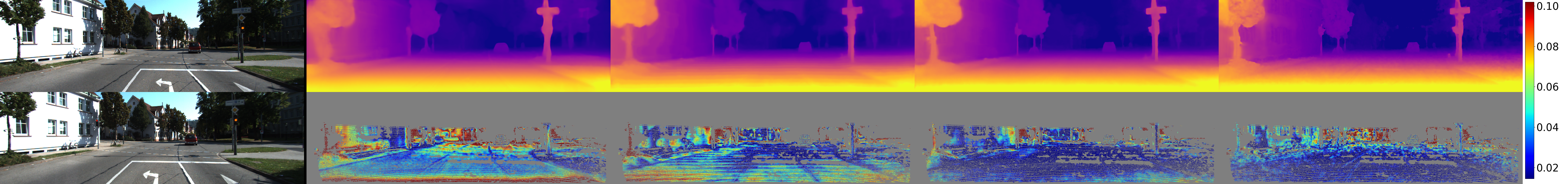}
  \includegraphics[width=0.97\textwidth]{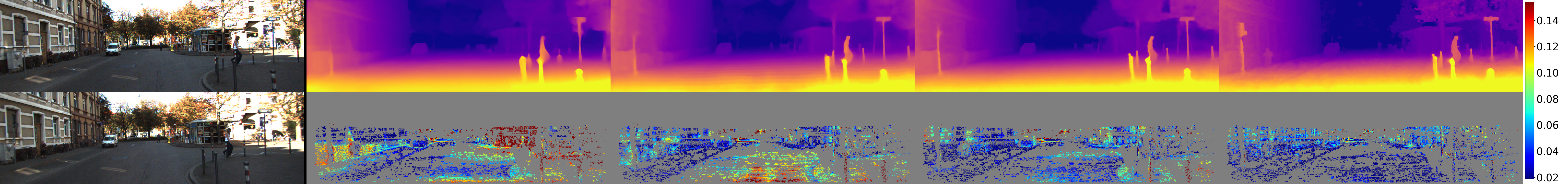}
  \includegraphics[width=0.97\textwidth]{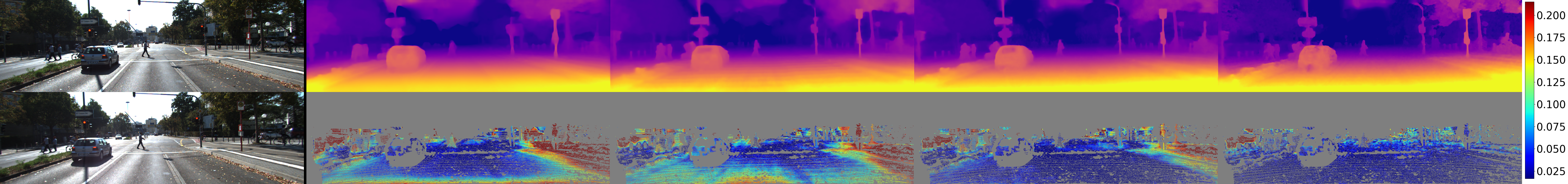}
  \includegraphics[width=0.97\textwidth]{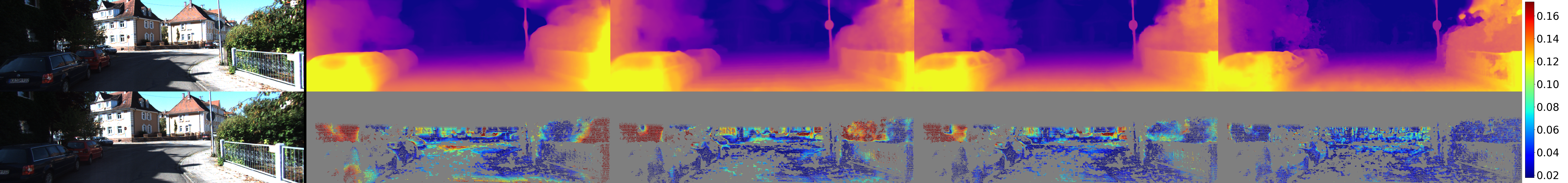}
  \leftline{\qquad \qquad Input Images
            \qquad \qquad \qquad \enspace EPCDepth~\cite{peng2021excavating}
            \qquad \qquad \quad SDFA-Net~\cite{zhou2022self}
            \qquad \qquad  TiO-Depth (Mono.)
            \qquad \qquad TiO-Depth (Bino.)}
  \end{center}
  \caption{Visualization results of EPCDepth~\cite{peng2021excavating}, SDFA-Net~\cite{zhou2022self} and our TiO-Depth on KITTI.
  The input stereo pairs are shown in the first column, where the left-view images are used for monocular depth estimation.
  The predicted depth maps with the corresponding `Abs. Rel.' error maps calculated on the improved Eigen test set are shown in the following columns.
  For the error maps, red indicates larger error, and blue indicates smaller error as shown in the color bars.}
\label{fig:comp2}
\end{figure*}

\section{Dataset and metric}
TiO-Depth is trained on the KITTI dataset~\cite{geiger2012we} and evaluated on the KITTI, Cityscapes~\cite{cordts2016cityscapes}, and DDAD~\cite{guizilini20203d} datasets as mentioned in Sec. 4 of the main paper.

In addition to the Eigen split~\cite{eigen2014depth} and the KITTI 2015 stereo benchmark~\cite{menze2015joint} which are employed for training and testing, an improved Eigen test set~\cite{uhrig2017sparsity} comprised of 652 images with high-quality depth labels is also used for evaluation.
The test set of Cityscapes~\cite{cordts2016cityscapes} which contains 1525 stereo pairs with the disparity maps provided by SGM~\cite{hirschmuller2005accurate} and the validation set of DDAD which contains 3950 single images and the aligned LiDAR depth labels are used for evaluating the cross-dataset generalization ability of TiO-Depth, 

The following seven metrics are used to evaluate the performances of monocular and binocular depth estimations on all the datasets:
\begin{itemize}
    \item Abs Rel: $\frac{1}{N}\sum_i{\frac{\left| \hat{D}_i - D^{gt}_i \right|}{D^{gt}_i}}$
    
    \item Sq Rel: $\frac{1}{N}\sum_i{\frac{\left| \hat{D}_i - D^{gt}_i \right|^2}{D^{gt}_i}}$
    
    \item RMSE: $\sqrt{\frac{1}{N}\sum_i{\left| \hat{D}_i - D^{gt}_i \right|^2}}$
    \item logRMSE: $\sqrt{\frac{1}{N}\sum_i{\left| \log\left(\hat{D}_i\right) - \log\left(D^{gt}_i\right) \right|^2}}$
    \item Threshold (A$j$):  $ \% \quad s.t. \quad \max{\left( \frac{\hat{D}_i}{D^{gt}_i}, \frac{D^{gt}_i}{\hat{D}_i} \right)}< a^{j}$
\end{itemize}
where $\{\hat{D}_i, D^{gt}_i\}$ are the predicted depth and the ground-truth depth at pixel $i$, and $N$ denotes the total number of the pixels with the ground truth.
In practice, we use $a^{j} = 1.25, 1.25^{2}, 1.25^{3}$, which are denoted as A1, A2, and A3 in all the tables.
EPE and D1 metrics are also adopted for the evaluation of binocular depth estimation as done in~\cite{liu2019unsupervised,liu2020flow2stereo}:
\begin{itemize}
  \item EPE: $\frac{1}{N}\sum_i{\left | \hat{d}_i - d^{gt}_i \right |}$
  \item D1: $ \% \quad s.t. \left (\left | \hat{d}_i - d^{gt}_i \right | > 3\right ) \vee  \left (\frac{\left | \hat{d}_i - d^{gt}_i \right |}{d^{gt}_i} > 0.05\right ) $
\end{itemize}
where $\{\hat{d}_i, d^{gt}_i\}$ are the predicted disparity and the ground-truth disparity at pixel $i$.

\begin{table*}[]
  \begin{center}
  \footnotesize
  \renewcommand\tabcolsep{1.5pt}
  \begin{tabular}{|l|cccc|ccc|cc|}
    \hline
    Methods       & Abs. Rel. $\downarrow$ & Sq. Rel. $\downarrow$ & RMSE $\downarrow$ & logRMSE $\downarrow$ & A1 $\uparrow$  & A2    $\uparrow$ & A3    $\uparrow$ & EPE $\downarrow$ & D1 $\downarrow$ \\\hline
    w. Cat module (321) & 0.069                  & 0.505                 & 3.442             & 0.123                & 0.947          & 0.983            & 0.992            & 2.074            & 15.952          \\
    w. Attn module (321)  & 0.053                  & 0.439                 & 3.214             & 0.106                & 0.965          & 0.987            & \textbf{0.994}            & 1.377            & 7.421           \\\hline
    w. MFM (1)         & 0.054                  & \textbf{0.423}                 & 3.211             & 0.109                & 0.960          & 0.986            & 0.993            & 1.483            & 8.784           \\
    w. MFM (21)       & 0.052                  & 0.445                 & 3.268             & 0.107                & 0.965          & 0.987            & \textbf{0.994}            & 1.305            & 7.077           \\
    TIO-Depth & \textbf{0.051}         & 0.429        & \textbf{3.137}    & \textbf{0.105}       & \textbf{0.966} & \textbf{0.988}   & \textbf{0.994}   & \textbf{1.281}   & \textbf{6.684}  \\\hline
    w/o. $L_{gui}$ & 0.053                  & 0.506                 & 3.378             & 0.108                & \textbf{0.966}           & 0.987           & 0.993           & 1.292          & 6.984          \\
    w/o. $L_{gui}$, $L_{cos}$  & 0.053                  & 0.522                 & 3.404             & 0.110                & 0.965           & 0.986           & 0.993           & 1.326          & 6.775          \\
    w/o. $L_{gui}$, $L_{cos}$, $M_{occ}$ & 0.054                  & 0.565                 & 3.637             & 0.121                & 0.963           & 0.984           & 0.992           & 1.345          & 7.159          \\\hline
  \end{tabular}
  \end{center}
  \caption{Binocular depth estimation results on KITTI 2015 training set in the ablation study (Tab. 4 in the main paper).
           The numbers in the name of methods mean the indexes of the used modules as shown in Fig. 2 of the main paper.
           All the results are evaluated after training 30 epochs.}
  \label{tab:abla-st2-12}
\end{table*}

\begin{table*}[]
  \begin{center}
  \footnotesize
  \renewcommand\tabcolsep{1.5pt}
  \begin{tabular}{|ccc|cccc|ccc|}
    \hline
    Steps & $L_{dis}$ & FB.         & Abs. Rel. $\downarrow$ & Sq. Rel. $\downarrow$ & RMSE $\downarrow$ & logRMSE $\downarrow$ & A1 $\uparrow$ & A2 $\uparrow$ & A3 $\uparrow$ \\\hline
    1     & -                & -          & 0.088                  & 0.556                 & 4.093             & 0.173                & 0.904           & 0.967           & 0.984           \\
    1+2   & -                & -          & 0.088                  & 0.557                 & 4.067             & 0.172                & 0.906           & 0.968           & 0.984           \\
    1+2+3 & $P^l_s$          & \checkmark & 0.086                  & 0.590                 & 4.021             & 0.169                & 0.911           & 0.969           & 0.985           \\
    1+2+3 & $P^l_h$          & \checkmark & \textbf{0.085}         & \textbf{0.544}        & \textbf{3.919}    & \textbf{0.169}       & \textbf{0.911}  & \textbf{0.969}  & \textbf{0.985} \\
    1+2+3 & $P^l_h$          &      -      & 0.098                  & 0.695                 & 4.367             & 0.183                & 0.892           & 0.964           & 0.983           \\\hline
    \end{tabular}
  \end{center}
  \caption{Monocular depth estimation results predicted by TiO-Depth on the KITTI Eigen test set in the ablation study (Tab. 5 in the main paper).
  'FB.' denotes using the final branches.}
  \label{tab:abla-st2}
\end{table*}

For the evaluation on the raw and improved KITTI Eigen test sets~\cite{eigen2014depth, uhrig2017sparsity}, we use the center crop proposed in~\cite{garg2016unsupervised} and the standard cap of 80m.
For the evaluation on the KITTI 2015 training set, all the ground truth disparities are used for calculating D1 and EPE metrics, while other metrics are calculated with the cap of 80m as done in~\cite{chen2021revealing}.
For the evaluation on the DDAD dataset~\cite{guizilini20203d}, the cap of 200m is used, while the input images are resized into the resolution of $384\times640$ as done in~\cite{guizilini20203d}.
For the evaluation on the Cityscapes dataset~\cite{cordts2016cityscapes}, we use the center crop and the standard cap of 80m as done in~\cite{watson2021temporal, gordon2019depth, li2021unsupervised}, while the input images are cropped and resized into the resolution of $192\times512$ as done in~\cite{watson2021temporal}.
All the cross-dataset results of TiO-Depth are calculated after the median scaling~\cite{eigen2015predicting}.

\begin{figure*}[]
  \begin{center}
  \footnotesize
  \begin{minipage}{0.55\textwidth}
    \includegraphics[width=\textwidth]{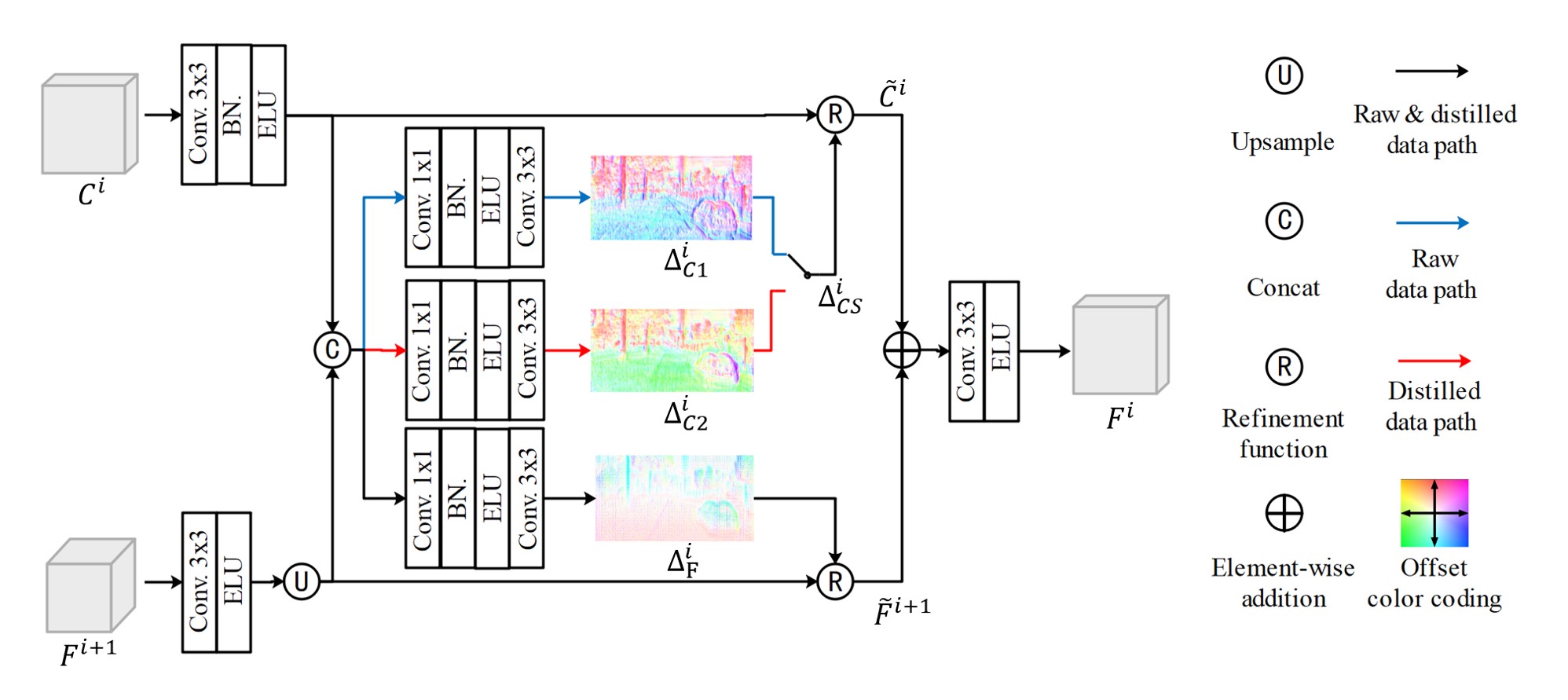}
    \centerline{(a)}
  \end{minipage}
  \begin{minipage}{0.018\textwidth}
    \includegraphics[width=\textwidth]{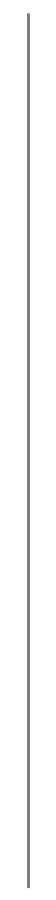}
  \end{minipage}
  \begin{minipage}{0.35\textwidth}
    \includegraphics[width=\textwidth]{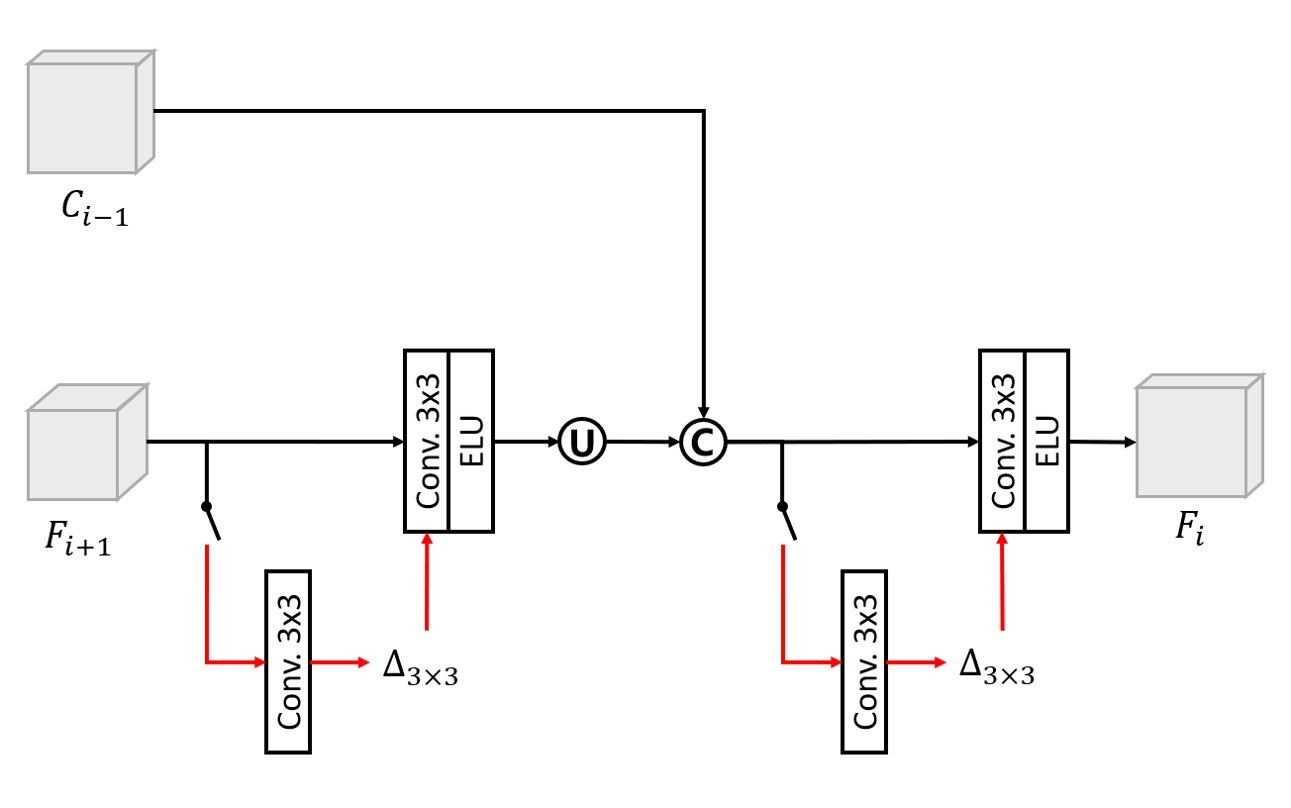}
    \centerline{(b)}
  \end{minipage}
  
  \end{center}
  \caption{(a) Architecture of the Self-Distilled Feature Aggregation (SDFA) block cited from~\cite{zhou2022self}.
           (b) Architecture of the switchable feature aggregation block inspired by the deformable convolution~\cite{dai2017deformable,zhu2019deformable}.}
\label{fig:module-arc}
\end{figure*}

\begin{table*}[]
  \begin{center}
  \footnotesize
  \renewcommand\tabcolsep{1.5pt}
  \begin{tabular}{|ccc|cccc|ccc|}
    \hline
    Steps & $L_{dis}$ & FB.         & Abs. Rel. $\downarrow$ & Sq. Rel. $\downarrow$ & RMSE $\downarrow$ & logRMSE $\downarrow$ & A1 $\uparrow$ & A2 $\uparrow$ & A3 $\uparrow$ \\\hline
    1     & -                & -          & 0.094                  & 0.579                 & 4.155             & 0.178                & 0.896           & 0.966           & 0.984           \\
    1+2   & -                & -          & 0.094                  & 0.582                 & 4.165             & 0.177                & 0.896           & 0.966           & 0.984           \\
    1+2+3 & $P^l_h$          & \checkmark & \textbf{0.086}         & \textbf{0.551}        & \textbf{3.967}    & \textbf{0.170}       & \textbf{0.907}  & \textbf{0.969}  & \textbf{0.985} \\
    1+2+3 & $P^l_h$          &   -         & 0.103                  & 0.688                 & 4.367             & 0.181                & 0.890           & 0.966           & 0.984           \\\hline
    \end{tabular}
  \end{center}
  \caption{Monocular depth estimation results predicted by the variant of TiO-Depth on the KITTI Eigen test set in the ablation study.}
  \label{tab:variant-abla-st}
\end{table*}

\begin{table*}[]
  \begin{center}
    \footnotesize
      \begin{tabular}{|lc|cccc|ccc|}
          \hline
          Method  &   Resolution  & Abs Rel$\downarrow$ & Sq Rel$\downarrow$ & RMSE$\downarrow$  & logRMSE$\downarrow$ & A1$\uparrow$    & A2$\uparrow$    & A3$\uparrow$    \\\hline
          TiO-Depth             & 192$\times$640 &  0.091   & 0.625  & 4.179 & 0.174  & 0.902 & 0.968 & 0.984 \\
          TiO-Depth             & 320$\times$1024  & 0.087   & 0.566  & 3.970 & 0.170  & 0.910 & 0.969 & 0.985  \\
          TiO-Depth             & 384$\times$1280 & 0.085 & 0.544 & 3.919 & 0.169 & 0.911 & 0.969 & 0.985 \\\hline
          TiO-Depth (Bino.)     & 192$\times$640  & 0.065   & 0.572  & 3.767 & 0.157  & 0.940 & 0.971 & 0.984 \\
          TiO-Depth (Bino.)     & 320$\times$1024  & 0.064   & 0.526  & 3.594 & 0.153  & 0.943 & 0.973 & 0.985 \\
          TiO-Depth (Bino.)    & 384$\times$1280   & 0.063 & 0.523 & 3.611 & 0.153 & 0.943 & 0.972 & 0.985 \\\hline
          \end{tabular}
  \end{center}
  \caption{Depth estimation results with different input image resolutions on the KITTI Eigen test set in the ablation study.}
  \label{tab:res}
\end{table*}

\section{Comparative Results}
\label{sec:comp}
As done in \cite{watson2019self, gonzalezbello2020forget, gonzalez2021plade, zhou2022learning, zhou2022self}, we evaluate TiO-Depth on the improved KITTI Eigen test set~\cite{uhrig2017sparsity} and the corresponding results are shown in~\cref{tab:main2}.
It can be seen that TiO-Depth outperforms all the comparative methods in most cases in both monocular and binocular (multi-frame) tasks.
Additional visualization results are given in~\cref{fig:comp}.
These results further demonstrate the effectiveness of TiO-Depth as a two-in-one model.

In~\cref{tab:gen}, the monocular and binocular depth estimation results of TiO-Depth and 6 comparison methods~\cite{li2021unsupervised,godard2019digging,guizilini20203d,guizilini2022multi,petrovai2022exploiting,watson2021temporal} on the DDAD~\cite{guizilini20203d} and Cityscapes~\cite{cordts2016cityscapes} datasets under all the seven metrics are given, which demonstrate the generalization ability of TiO-Depth on the unseen datasets.

\section{Ablation Study}
We have verified the effectiveness of each key element in TiO-Depth by conducting ablation studies on the KITTI dataset~\cite{geiger2012we} in Sec. 4.3 of the main paper.
\cref{tab:abla-st2-12} shows the binocular depth estimation results in the ablation study under all of the nine metrics, which demonstrate the effectiveness of the dual-path decoder and the stereo loss $L_S$ on the binocular task.

The monocular depth estimation results in the ablation study under all of the seven metrics are shown in \cref{tab:abla-st2}, which indicate the effectiveness of the multi-stage joint-training strategy.
Furthermore, the results also prove the significance of the final branches in the Self-Distilled Feature Aggregation (SDFA)~\cite{zhou2022self} blocks (as shown in~\cref{fig:module-arc}(a) where the raw data path in blue is used as the auxiliary branch and the distilled branch in red is used as the final branch) for the monocular task.

To further explore the effect of such switchable branches on learning more accurate monocular depths, a variant of TiO-Depth is built by replacing the three SDFA blocks in the dual-path decoder by the switchable aggregation blocks shown in~\cref{fig:module-arc}(b).
The switchable aggregation block is inspired by the deformable convolution~\cite{dai2017deformable,zhu2019deformable} and is built based on the basic decoder block described in Sec. 3.2 of the main paper.
In comparison to the basic decoder block, it employs two additional $3\times3$ convolutional layers as the switchable `final branches' to learn the spatial offsets for the kernels of the convolutional layers in the basic decoder block.
Accordingly, the standard convolutional layers in the basic block are converted to the deformable convolutions when the final branches are used.
We train this variant with the multi-stage joint-training strategy and conduct the ablation studies.
The corresponding results are shown in~\cref{tab:variant-abla-st}.
It can be seen that the whole performances of the variant TiO-Depth are poorer than that of TiO-Depth shown in~\cref{tab:abla-st}, mainly because the SDFA blocks could aggregate the features more effectively than the basic decoder layers.
However, using the switchable final branches significantly improves the performance of the model in comparison to that without the final branches.
These results further demonstrate that the potential of TiO-Depth for employing a more general architecture.

Finally, we conduct the ablation study on the input image resolution. As seen from~\cref{tab:res}, TiO-Depth still performs well under the two low resolutions.

\end{document}